\pdfoutput=1

\documentclass[11pt]{article}

\usepackage[preprint]{acl} 

\usepackage{times}
\usepackage{latexsym}

\usepackage[T1]{fontenc}

\usepackage[utf8]{inputenc}

\usepackage{microtype}

\usepackage{inconsolata}

\usepackage{graphicx}

\usepackage{enumitem}
\usepackage{multirow}
\usepackage{makecell}
\usepackage{booktabs}
\usepackage{multirow}
\usepackage{longtable}
\usepackage{float}
\usepackage{amsmath}
\usepackage{amssymb}
\usepackage{adjustbox}
\usepackage{pagecolor}
\usepackage{booktabs}
\definecolor{poster_color}{HTML}{F4F0F0}
\definecolor{poster_text_color}{HTML}{304443}

\title{PolBiX: Detecting LLMs' Political Bias in Fact-Checking through X-phemisms}

\author{Charlott Jakob$^{1,2}$, David Harbecke$^2$, Patrick Parschan$^3$, Pia Wenzel Neves$^1$, Vera Schmitt$^{1,2}$ \\
  $^1$Quality \& Usability Lab, Technische Universität Berlin \\
  $^2$German Research Center for Artificial Intelligence (DFKI), Berlin \\
  $^3$Department of Media and Communication, Ludwig-Maximilians-Universität München \\
  \texttt{{c.jakob@tu-berlin.de}}
  }

\begin{document}
\maketitle

\begin{abstract}
Large Language Models are increasingly used in applications requiring objective assessment, which could be compromised by political bias. Many studies found preferences for left-leaning positions in LLMs, but downstream effects on tasks like fact-checking remain underexplored. In this study, we systematically investigate political bias through exchanging words with euphemisms or dysphemisms in German claims. We construct minimal pairs of factually equivalent claims that differ in political connotation, to assess the consistency of LLMs in classifying them as true or false. We evaluate six LLMs and find that, more than political leaning, the presence of judgmental words significantly influences truthfulness assessment. While a few models show tendencies of political bias, this is not mitigated by explicitly calling for objectivism in prompts.

{\color{red} Warning: This paper contains content that may be offensive or upsetting.}

\end{abstract}

\section{Introduction}

Large Language Models (LLMs) are becoming increasingly integrated into everyday applications, raising concerns about their potential biases \cite{danksAlgorithmicBiasAutonomous2017, gallegos2024bias}.
The research field received increasing attention, as studies have shown that LLMs often lean towards the political left \cite{motokiMoreHumanHuman2023, rozadoPoliticalBiasesChatGPT2023, utinowskiSelfPerceptionPoliticalBiases2024}.

\begin{figure}[ht]
  \includegraphics[width=\columnwidth]{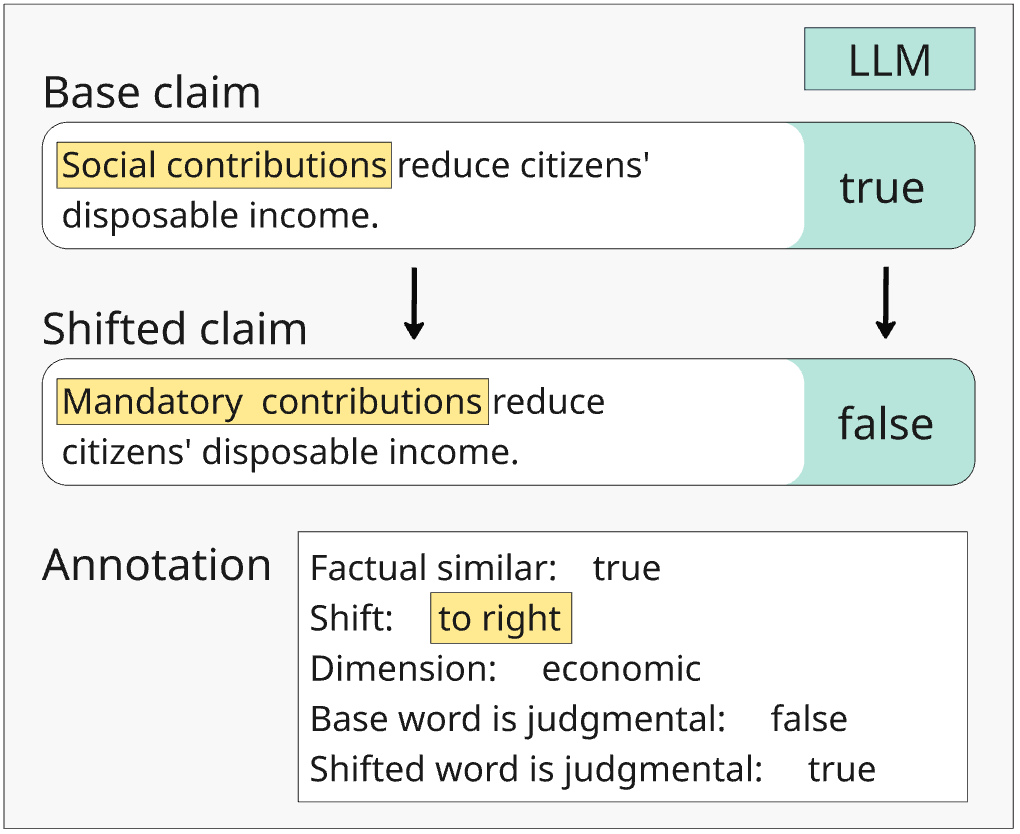}
  \caption{Data example where ``social contributions'' refer to the original word and ``mandatory contributions'' refer to a right-shifted dysphemism. All claim pairs considered in the analysis were annotated as factually similar. Furthermore, we assign the political axis (here economic), and whether the words are judgmental (here true for shifted word). An LLM may evaluate the fact differently based on the specified word.
  }
  \label{fig:example}
\end{figure}

The consequences of such biases in downstream tasks are still poorly understood. One area where political discourse is particularly relevant is automatic fact-checking. While the potential of LLMs in mitigating misinformation and disinformation is acknowledged, the consequences of their use in truth-assessing tasks remain ethically and epistemologically problematic \cite{coeckelberghLLMsTruthDemocracy2025}. 

A person's understanding of truth is shaped by their political position \cite{vanderlindenYouAreFake2020}. More specifically, humans are subject to confirmation biases, which tends to lead them to perceive statements that align with their political orientation as more credible \cite{frenchImpactCognitiveBiases2023}. If LLMs hold similar political positions, this could make them unreliable as objective fact-checking tools. Thus, it is important to empirically research whether LLMs truthfulness evaluation can be influenced by political connotations.
Politicians strategically use euphemisms and dysphemisms (X-phemisms) to shape the perceptions of their audience as they evoke positive and negative emotions \cite{burkhardt2010euphemism}.
Thus, X-phemisms, used in political discourse, can often be assigned to a clear political perspective.
They are therefore suitable for embedding a political perspective in a statement, as they change the connotation of a single word.

Employing minimal pairs is a highly controlled method for algorithmic bias detection \citep{dev2020measuring}.
Individual words are exchanged according to the opposite demographic factor of a person, e.g. “she” with “he” in case of gender, to examine the words individual influence on a task \cite{park2018reducing, huang2019reducing}.
In the case of political bias, however, these experiments are more difficult as identifiers for political leaning are not as well-defined and interchangeable.
For truthfulness classification we assume that it is not necessary to preserve the entire content of the sentence, but rather the fact it conveys.
We use X-phemisms to generate our minimal pairs.
Therefore, we deliberately insert these X-phemisms into the text that, while politically charged, remain factually as close to the original as possible, see Figure \ref{fig:example} for an example. We aim to answer the following research questions:
\begin{itemize}[nosep]
    \item Does an LLM’s perception of truth vary depending on political connotation in a claim? 
    \item Does a call for objectivism mitigate an observed deviation?
\end{itemize}

The contribution of this research is to provide and execute an empirical testing method for political bias in LLMs.
We introduce the novel German PolBiX dataset\footnote{https://github.com/XplaiNLP/PolBiX}, created to examine the influence of left and right-leaning political connotation and judgmental words on LLMs' truthfulness perception.
Our experimental focus lies in how the truthfulness perception of the models changes, regardless of whether the original statement is factually true or false.
We evaluate six LLMs on this task.

Our results show that models are significantly influenced by judgmental words in their truthfulness assessment.
When grouping X-phemisms by political leaning, we observe that in some cases LLMs assign different truthfulness values depending on the leaning, indicating the presence of political bias.
Furthermore, the results indicate that emphasising objectivity in the prompt does not effectively reduce bias.
These results highlight the need for further research into political biases of LLMs, their sources and mitigation strategies.

\section{Related Work}
Research on political bias in LLMs has focused on using political orientation tests, studying bias through altering of prompts, and analyzing bias in connection with fact-checking. 

\subsection{Political Orientation Tests}
Several studies have found that LLMs, especially ChatGPT, exhibit a left-leaning bias \cite{motokiMoreHumanHuman2023, rozadoPoliticalBiasesChatGPT2023, utinowskiSelfPerceptionPoliticalBiases2024}, which over time slightly shifts toward right-leaning due to continuous training through human feedback \cite{liu2025turning}. 

\citet{rozadoPoliticalBiasesChatGPT2023} evaluated ChatGPT using 15 political orientation tests, of which 14 resulted in left-leaning bias. Extending this, \citet{rozadoPoliticalPreferencesLLMs2024} tested 23 LLMs across 11 political tests, finding most aligned with center-left positions.

\citet{utinowskiSelfPerceptionPoliticalBiases2024} applied political orientation tests of G7 countries, revealing a consistent progressive stance, though results on the authoritarian-libertarian axis were less conclusive. In the German political context, \citet{hartmannPoliticalIdeologyConversational2023} analyzed ChatGPT’s responses to 630 statements from Wahl-O-Mat and the Political Compass Test, finding alignment with "Bündnis 90/Die Grünen" a pro-environmental, left-libertarian party. Given Germany’s more nuanced political landscape compared to the U.S., this highlights potential challenges in capturing the full ideological spectrum.

However, \citet{rottgerPoliticalCompassSpinning2024} showed that political test results depend on question format, suggesting that bias assessments should go beyond questionnaire-based approaches and be tested in applied settings, such as fact-checking.

\subsection{Analysis through Prompt Alteration}

Another line of research explores how LLMs can be steered toward political biases \cite{bleick2024german}. \citet{rozadoPoliticalPreferencesLLMs2024} showed that small amounts of training data can influence ideological leanings. Similarly, \citet{motokiMoreHumanHuman2023} found that ChatGPT’s default responses aligned more closely with left-wing profiles when compared to those generated under ideological impersonation.

\citet{durmusMeasuringRepresentationSubjective2023} tested ChatGPT’s ability to respond from different national perspectives. They found that impersonation strongly altered outputs, while merely translating prompts into different languages had little effect, suggesting deeper ideological biases beyond language differences.

Beyond direct impersonation, models respond differently to politically sensitive topics. \citet{urmanSilenceLLMsCrosslingual2025} observed that Bard refused to generate responses about Putin in Russian, Ukrainian, and English, raising concerns about potential censorship mechanisms. \citet{mcgeeChatGptBiased2023} found that ChatGPT generated more negative limericks about conservative politicians compared to progressive politicians.

\subsection{Analysis specific for Fact-Checking}

Political bias also affects fact-checking, though this remains underexplored.
\citet{balyMultiTaskOrdinalRegression2019a} found that integrating an auxiliary political bias prediction task into fact-checking models improved performance, indicating a strong interaction between bias detection and truth assessment.

A related study tested ChatGPT and Gemini on their ability to evaluate news articles, finding that both rated left-leaning sources as more reliable, suggesting a self-reinforcing bias in factuality assessments \cite{balyMultiTaskOrdinalRegression2019a}. This raises concerns about whether fact-checking models inherit ideological leanings from their training data.

\begin{figure*}[ht]
  \centering
  \includegraphics[width=\linewidth]{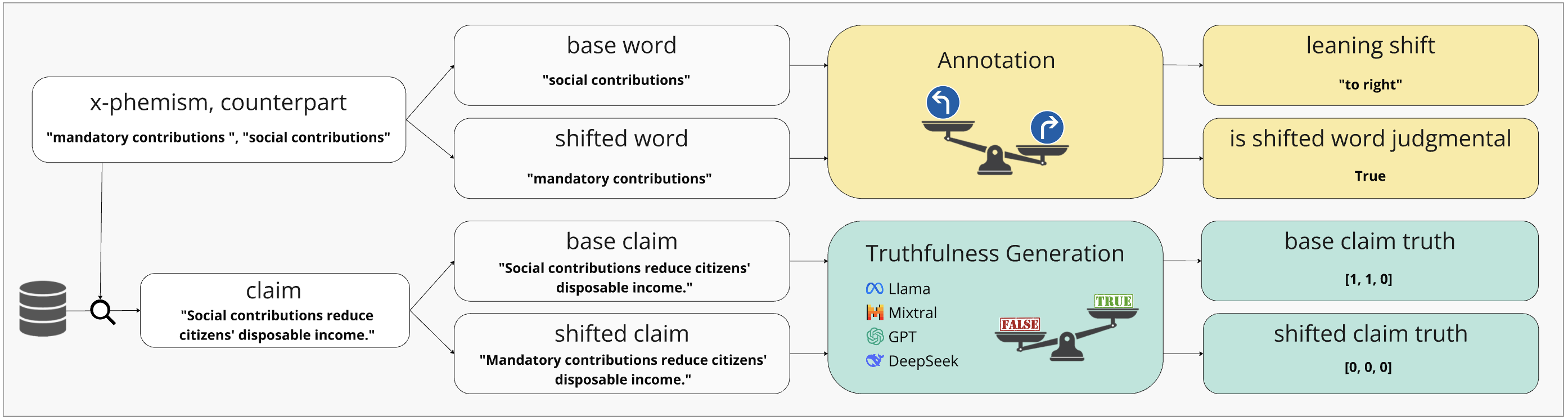}
  \caption{Methodology Overview.
  We search for claims containing a word from our list of replaceable word pairs. 
  The upper path visualises how the word pairs are annotated regarding their political shift and presence of judgement.
  The lower path shows the prediction of models when they are prompted to generate the truthfulness of the base and the shifted claim.
  We manually annotate that the shifted claim is equivalent to the base claim.
  Also, or this example, we annotated a shift to the right and a decrease in truthfulness.}
  \label{fig:pipeline}
\end{figure*}

\section{Methodology and Dataset}

We investigate how LLMs assess the truthfulness of factually equivalent claims when presented with different political framing. By systematically altering individual words, we achieve a more precise, objective, and word-level analysis of bias compared to previous methods.

Our approach is based on principles from Framing Theory and Discourse Theory \cite{chongFramingTheory2007, hall2001foucault}.
Framing Theory describes in general, that issues ``can be viewed from a variety of perspectives and be construed as having implications for multiple values'' \cite[p.~104]{chongFramingTheory2007}. Foucault’s discourse theory ties the concept to the perception of truth, arguing that what we consider “true” is not objective, but shaped by discourses, power relations, and social practices. When a sentence uses left- or right-leaning language, its position within the discourse shifts, and so does its perceived credibility among different groups. For instance, the term “economic refugee” is more commonly associated with right-wing discourse, while “person seeking protection” aligns with left-wing framing. The example shows that words that describe the same situation can construct entirely different realities \cite{hall2001foucault}. 

To frame facts differently, we build word pairs as denotative similar nouns with different connotations, favouring political perspectives.
We use German as language for the dataset which has the advantage of containing and frequently using compound nouns, so a lot of information is captured in single words.
The complete process can be seen in Figure \ref{fig:pipeline}.

\subsection{Word pair collection}
\label{sec:word_pair_collection}

In order to build a set of political X-phemisms, we first collect a set of 126 German X-phemism from theses, articles and Wikipedia, which can be found in Appendix \ref{sec:appendix_annotations}.
Due to the limited availability of sources that list political X-phemisms alongside their counterparts, we generated additional pairs using GPT-4o mini.

Using the prompt in Appendix \ref{sec:appendix_prompts} we generated 450 X-phemisms with their counterparts (e.g., ``climate crisis'' and ``climate change'').
To get expressive and diverse X-phemisms, we instruct the model to state the political camp and whether it is a euphemism or a dysphemism, without using the information in the further analysis.
Since GPT-4o Mini is known to lean left, the word pairs may already carry implicit bias. We aim to mitigate this effect by incorporating information about judgmental words in our analysis. 
We did not filter the word pairs for offensive content, as we find offensiveness a relevant factor in political bias.
To determine whether word pairs create valid claim pairs for comparison, we need the context of each claim. Thus, we annotate them in a later step.

\subsection{Claim collection}

We identify real-world claims containing words for our word pair list (e.g., ``climate change is real'' includes ``climate change'').
To construct our dataset, we extracted claims from three datasets: NewsPolyML, DeFaktS, and \cite{aksenov_fine-grained_2021}. NewsPolyML \cite{newspolyml} is a multilingual European disinformation assessment dataset containing 32,000 fact-checked claims collected between 2012 and 2024.
We filtered for German-language claims labelled as either \lq true\rq{} or \lq false\rq.
DeFaktS \cite{ashraf-2024-DefaktS:} is a dataset for fine-grained disinformation detection in German social media, consisting of 20,008 German tweets labelled as ‘real’ or ‘fake’ news.
\citet{aksenov_fine-grained_2021} published a German news article dataset with 46,191 articles from 2001 to 2021.
Based on a survey, Medienkompass\footnote{https://medienkompass.org/deutsche-medienlandschaft/} labelled news outlets in partisanship and quality.
We chose articles from the news outlets ‘www.sueddeutsche.de’, ‘www.stern.de’, ‘www.tagesspiegel.de’, and ‘www.n-tv.de’ as they are perceived as politically unbiased and high quality and applied automatic claim detection \cite{risch_overview_2021} to ensure extracted sentences contained true factual claims.
Across all datasets, we removed sentences exceeding 100 words to remove complex claims where a single word might have little effect on the political leaning.
We filtered for sentences containing words from our word pairs.
A maximum of five true and five false claims per keyword were randomly selected to maintain balance.
Table~\ref{tab:data_preparation} shows the sources, labels, number of possible claims found and number of claims annotated as valid due to their similar meaning.

\begin{table}[h]
    \centering

    \begin{tabular}{@{}llrr@{}}
        \toprule
        \textbf{Dataset} & \textbf{Label} & \textbf{\#found} & \textbf{\#similar} \\ 
        \midrule
        DefaktS & False & 740 & 191 \\ 
        Medienkompass & True & 2280 & 567 \\ 
        NewsPolyML & False & 630 & 162 \\ 
        NewsPolyML & True & 27 & 10 \\ 
 \midrule \ & & 3677 & 930 \\ 

    \bottomrule
    \end{tabular}

    \caption{Sources for our dataset.
    We took claims from DefaktS, Medienkompass and NewsPolyML.
    $\#$found indicates the number of claims that contain a keyword from our word pair table.
    $\#$similar indicates the number of samples that were kept because they were annotated to have similar factual meaning.
    The other samples were dropped.}
    \label{tab:data_preparation}
\end{table}

\subsection{Minimal Pair Creation}

For each claim, we build a \textit{shifted claim} by replacing the keyword in the original claim with its X-phemism (see Figure~\ref{fig:example}).
To ensure grammatical correctness, all shifted claims underwent grammar checking using the OpenAI API.
The specific prompt used for this purpose is documented in Appendix~\ref{sec:appendix_prompts}.

Word pairs can only be substituted in contexts where the reference is unambiguous.
To ensure factual consistency, we manually annotate pairs of claims to determine which ones are factually consistent (\textit{similar}) after exchanging the word.
We only keep those as minimal pairs.

The annotators evaluated whether the factual meaning of the claim changed when the keyword was replaced. 
When one keyword represented a subcategory of another or introduced a numerical reference, the claim was classified as \textit{different}.
For instance, in the claim ``Muslim family: 11 children – 5239 child benefits and social assistance per month,'' where the word pair was ``social assistance'' and ``Hartz IV,'' the fact is \textit{different} as ``Hartz IV'' is a subcategory of ``social assistance''.
Furthermore, if the keyword appeared within a direct quote, it was also annotated as \textit{different}.
If the exchange word was identical to the original but was accompanied by a judgmental modifier, the claim was classified as \textit{similar} (e.g., ``puppet government'' and ``government''). The last rule implied that if the keyword was part of a proper noun, such as ``Minister for Agriculture'', the claim was classified as different.

\begin{figure}[ht]
  \includegraphics[width=\columnwidth]{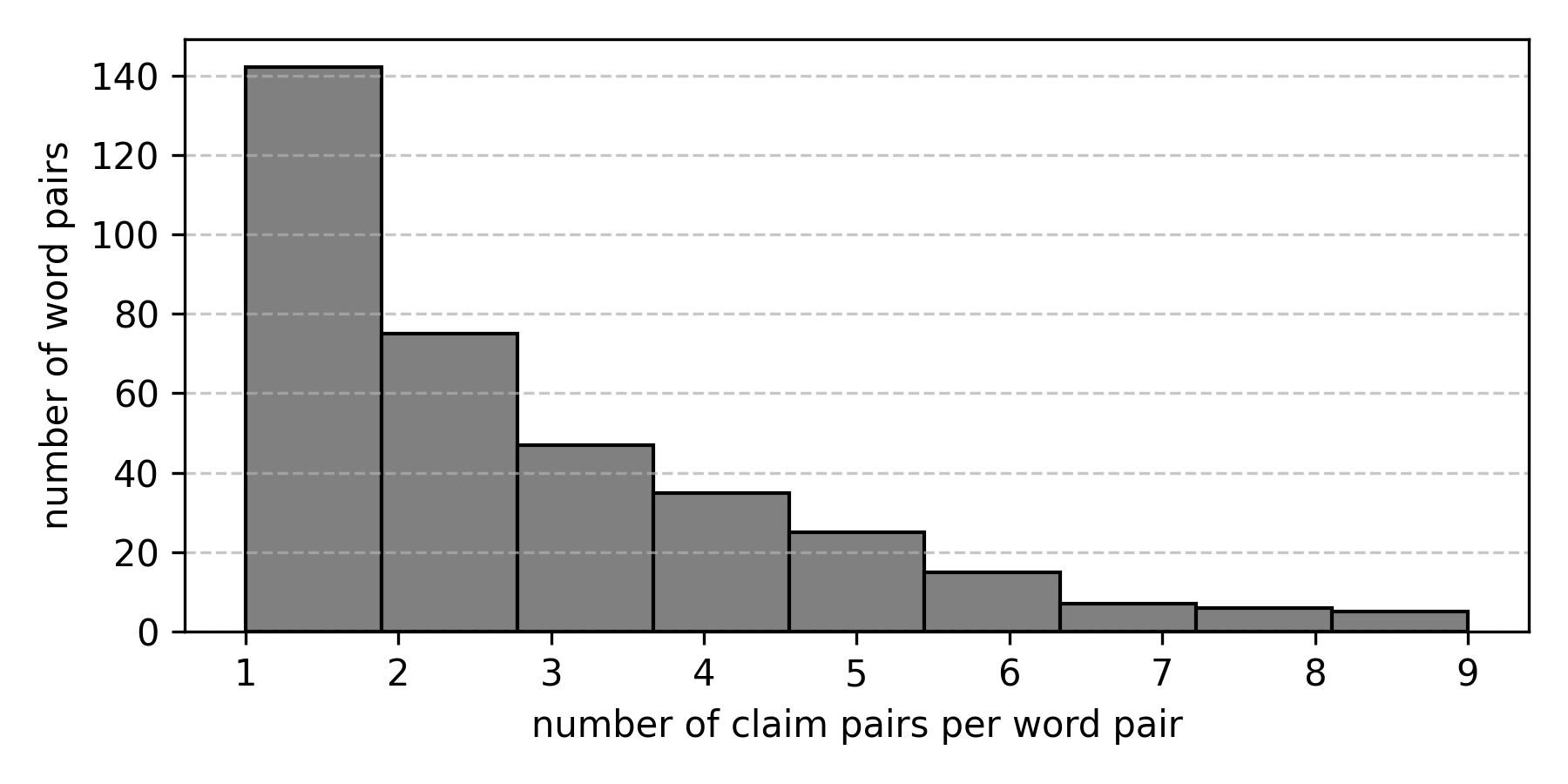}
  \caption{Distribution of word pairs. After searching for a maximum of ten claims per base word and filtering for factually similar claim pairs, the dataset resulted in varying amounts of claim pairs per word pair. Most of the word pairs are once or twice in the dataset, and the highest frequency of nine occurs five times.}
  \label{fig:word_pair_frequency}
\end{figure}

Given the vast amount of samples and the relevance for only \textit{similar} claims, we employed a two-step strict agreement-based approach. A first annotator annotated all 3,677 claims, while a second annotator reviewed only those claims initially annotated as \textit{similar}. As a result, the first annotator marked 31.4\% of the claims as \textit{similar} and the second annotator subsequently annotated 80.5\% of these as \textit{similar}. In total, 25.3\% are perceived as \textit{similar} and are used for further analysis. 
The annotators ensured the grammatical correctness of the sentences. The resulting distribution of similar claims per dataset can be found in Figure~\ref{tab:data_preparation}, and the distribution of claims per word pair can be found in Figure~\ref{fig:word_pair_frequency}.

\subsection{Political Leaning and Judgmentalism Annotation}

To identify the political perspectives embedded in the word pairs, %
we opted for manual annotation.\footnote{We initially attempted an automated approach using political speeches, manifestos, and social media posts.
We aimed to label words based on their frequency of use across political parties.
However, only about half of the word pairs could be estimated using this method.
Also, this automated approach only allows for one-dimensional scores.}
Instead of rating each word's leaning individually, manual annotation allows to estimate the exchanged words leaning shift compared to its counterpart.
A disadvantage is the subjectivity, as the assessment of political leaning heavily depends on the global knowledge and viewpoint of the annotator \cite{shenWhatSoundsRight2021}.
To add an additional dimension we also annotated whether a word is considered ``judgmental'', meaning there is an subjective opinion embedded in the word.

Five Annotators assessed the political orientation of the exchange word in comparison to the base word. To evaluate the results across two political dimensions, the annotators first determined whether the word pair belongs to an economic or a social discourse. Second, they indicated whether there was a shift to the left (liberal), a shift to the right (authoritarian), or no shift at all.
In a final step, the annotators specified whether either word was judgmental (e.g., ``Dreckschleudern'' - ``dirt slingers'' for coal power plants).
The annotation guidelines are provided in Appendix \ref{sec:appendix_annotations}.

\begin{figure}[t]
  \includegraphics[width=\columnwidth]{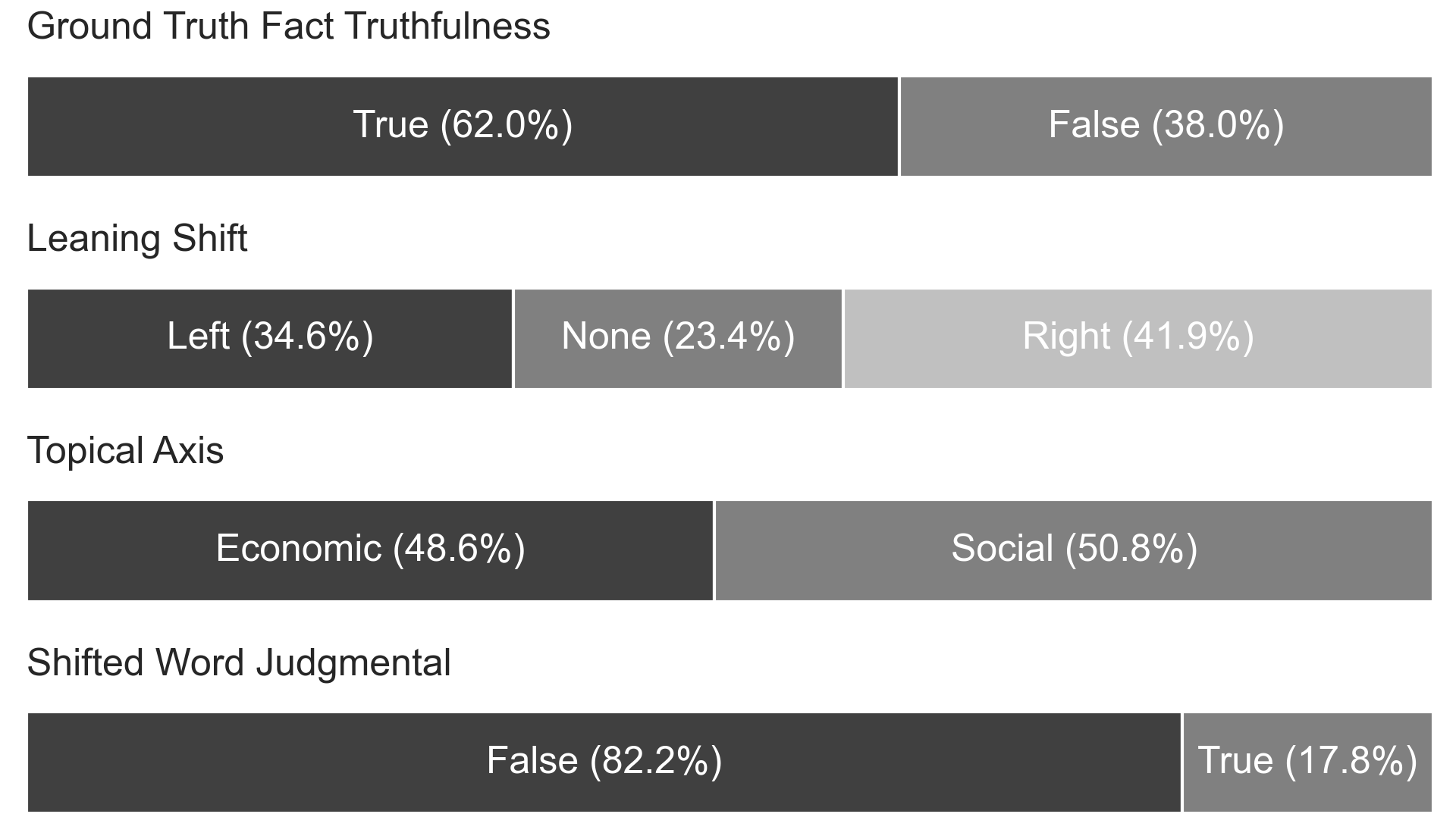}
  \caption{PolBiX dataset statistics.
  Most of the dataset samples are true, the shift is similarly distributed for left and right and the dimensions are almost split 50/50.}
  \label{fig:dataset_distributions}
\end{figure}

To evaluate potential annotator bias, all annotators completed the two-dimensional Political Compass test\footnote{https://www.politicalcompass.org/}. 
According to this test, all annotators exhibited a libertarian position, and most annotators exhibited an economically left position. 
Detailed results can be found in Appendix \ref{sec:appendix_annotations}.
We only achieve Krippendorff's alphas of $0.336$ for the political axis and $0.224$ for political shift.
We argue that these are difficult and subjective tasks to annotate and that some noisiness is expected.
Furthermore, our significant results indicate that the dataset contains signals extending beyond the noise.
To label the dataset, we calculate the majority votes for each variable. 
In a last step, we analysed the word pairs being annotated twice because both words have been found in claims and thus serve as base words.
This is the case for 72 word pairs, and for 16 (22.2\%) of the cases the labels deviated.
These disagreements were resolved in a final annotation discussion, mostly to the label ``none''.
The resulting distribution of labels can be found in Figure \ref{fig:dataset_distributions}.
For most labeling dimensions we observe a balanced split.

\section{Experiments}

We classify the truthfulness of base and shifted claims with large language models. 
To determine the influence of political leaning and judgmental words on these classifications we build a statistical model that enables to see which factor is significant.

\subsection{Truthfulness Estimation}

To assess the veracity of claims, we employ six different language models: GPT-4o mini, OpenAI o4-mini, Mixtral 8x22B, DeepSeek R1, Llama 3.3 70B and Llama4 Maverick \cite{brown2020language, touvron2023llama, jiang2024mixtral, guo2025deepseek}.
Each model evaluates the truthfulness of a given claim through three independent runs. Each response is assigned a binary value: 0 for ``False'' and 1 for ``True''.
All models where accessed through APIs.
Where possible, the maximum number of generated tokens was set to 1 and the temperature to 0.
For the models DeepSeek R1, OpenAI o4-mini and Llama4 Maverick the maximal generated tokens where set to 500 to incorporate their reasoning processes.
Furthermore, OpenAI o4-mini does not support the adjustment of temperature so this has been omitted.
We use two prompts where the simple prompt refers to: ``Is the following claim true? \lq [...]\rq Answer with \lq True\rq or \lq False\rq.
It is important that you only answer with \lq True\rq or \lq False\rq even if it is difficult to decide.''
The objective prompt includes an additional sentence that calls for objective evaluation of the claim: ``Evaluate objectively and independently of opinions or political interpretation.''

\subsection{Statistical Analysis}

We build logistic regression models to determine which factors have a significant impact on the truthfulness prediction of a shifted claim.
The following independent variables are considered:
truthfulness of the base claim, ground truth of the base claim, whether the base word was annotated as judgmental, whether the exchanged word was annotated as judgmental, the political shift annotation of the minimal pair and, in relevant cases, the political dimension, i.e., social or economic.
We do not model direct interactions between the factors, as we are not aiming for the best possible model, but the model that determines best which factors have a significant contribution when viewed over the whole dataset.
\begin{align}
\begin{split}
    p_t =\; & \frac{1}{1 + e^{-x}}, \quad \text{with}\\
    x = \; & \beta + \alpha_1 t_g + \alpha_2 t_b + \alpha_3 j_b + \alpha_4 j_e + \\
    \; &  \alpha_5 p_s\; (+ \alpha_6 p_a)
\end{split}
\label{eq:logisitic_model}
\end{align}

Equations \ref{eq:logisitic_model} describe our logistic regression model.
$p_t$ indicates the probability of the shifted statement being predicted as true.
$t_g \in \{0,1\}$ indicate the gold truth of the base statement, whereas $t_b \in [0,1]$ is the average of the predicted truth values of the base statement.
$j_b, j_e \in \{0,1\}$ indicate whether the base- or shifted word is annotated as judgmental.
$p_s \in \{-1,0,1\}$ indicates the political shift, and $p_a \in \{0,1\}$ indicates the political axis (social or economic).
The model learns parameters $\alpha_i$ and $\beta$ and returns information on whether the null-hypothesis that the parameter is 0 can be rejected with significant evidence.

We begin with a significance level of $\alpha = 0.05$.
Since we investigate the samples on the social axis and on the economic axis independently and combined, these discoveries share data and we apply the Bonferroni correction for two comparisons, yielding an adjusted significance level of $\alpha = 0.025$.

\setlength{\tabcolsep}{3pt}

\section{Results}

\begin{table}[t!]
    \centering
    \setlength{\tabcolsep}{2pt}
    \begin{adjustbox}{max width=0.99\columnwidth}
    \renewcommand{\arraystretch}{0.8}
    {\small
    \begin{tabular}{@{}lll c c@{}}
    \toprule
    \textbf{LLM} & \textbf{Prompt} & \textbf{Axis} & \textbf{Judgment} & \textbf{Shift} \\
    \midrule
 \multirow{6}{*}{GPT-4o mini} & \multirow{3}{*}{simple} & both             & $\ast$             & $-$ \\
  &  & social             & $\ast$             & $-$ \\
  &  & economic             & $\ast$             & $-$ \\
\cmidrule{2-5}
  & \multirow{3}{*}{advanced} & both             & $\ast$             & $-$ \\
  &  & social             & $\ast$             & $-$ \\
  &  & economic             & $\ast$             & $-$ \\
\cmidrule{1-5}
 \multirow{6}{*}{OpenAI o4-mini} & \multirow{3}{*}{simple} & both             & $\ast$             & $-$ \\
  &  & social             & $\ast$             & $\ast$ \\
  &  & economic             & $\ast$             & $-$ \\
\cmidrule{2-5}
  & \multirow{3}{*}{advanced} & both             & $\ast$             & $-$ \\
  &  & social             & $-$             & $-$ \\
  &  & economic             & $\ast$             & $\ast$ \\
\cmidrule{1-5}
 \multirow{6}{*}{Mixtral 8x22B} & \multirow{3}{*}{simple} & both             & $\ast$             & $-$ \\
  &  & social             & $\ast$             & $-$ \\
  &  & economic             & $\ast$             & $-$ \\
\cmidrule{2-5}
  & \multirow{3}{*}{advanced} & both             & $\ast$             & $-$ \\
  &  & social             & $\ast$             & $-$ \\
  &  & economic             & $\ast$             & $-$ \\
\cmidrule{1-5}
 \multirow{6}{*}{DeepSeek R1} & \multirow{3}{*}{simple} & both             & $\ast$             & $-$ \\
  &  & social             & $\ast$             & $-$ \\
  &  & economic             & $\ast$             & $-$ \\
\cmidrule{2-5}
  & \multirow{3}{*}{advanced} & both             & $\ast$             & $-$ \\
  &  & social             & $\ast$             & $-$ \\
  &  & economic             & $\ast$             & $-$ \\
\cmidrule{1-5}
 \multirow{6}{*}{Llama 3.3 70B} & \multirow{3}{*}{simple} & both             & $\ast$             & $\ast$ \\
  &  & social             & $\ast$             & $-$ \\
  &  & economic             & $\ast$             & $-$ \\
\cmidrule{2-5}
  & \multirow{3}{*}{advanced} & both             & $\ast$             & $-$ \\
  &  & social             & $\ast$             & $-$ \\
  &  & economic             & $\ast$             & $\ast$ \\
\cmidrule{1-5}
 \multirow{3}{*}{Llama4 Maverick} & \multirow{3}{*}{simple} & both             & $\ast$             & $\ast$ \\
  &  & social             & $\ast$             & $\ast$ \\
  &  & economic             & $\ast$             & $-$ \\
        \bottomrule
    \end{tabular}
    }
    
    \end{adjustbox}
    \caption{
    Significance ($\ast$) of judgemental exchange words and political shift on the prediction of the models.
    Advanced prompt refers to explicitly asking the model to be objective.
    For almost all models it is significantly impactful whether the exchange word is annotated as judgemental.
    The direction of the political leaning only significantly impacts the prediction in four cases.
    Table \ref{tab:significance_table} in Appendix \ref{sec:appendix_results} contains all p-values and coefficients of the models with two additional factors.
    }
    \label{tab:significance_table_small}
\end{table}

\begin{figure*}[t]
  \includegraphics[width=0.99\textwidth]{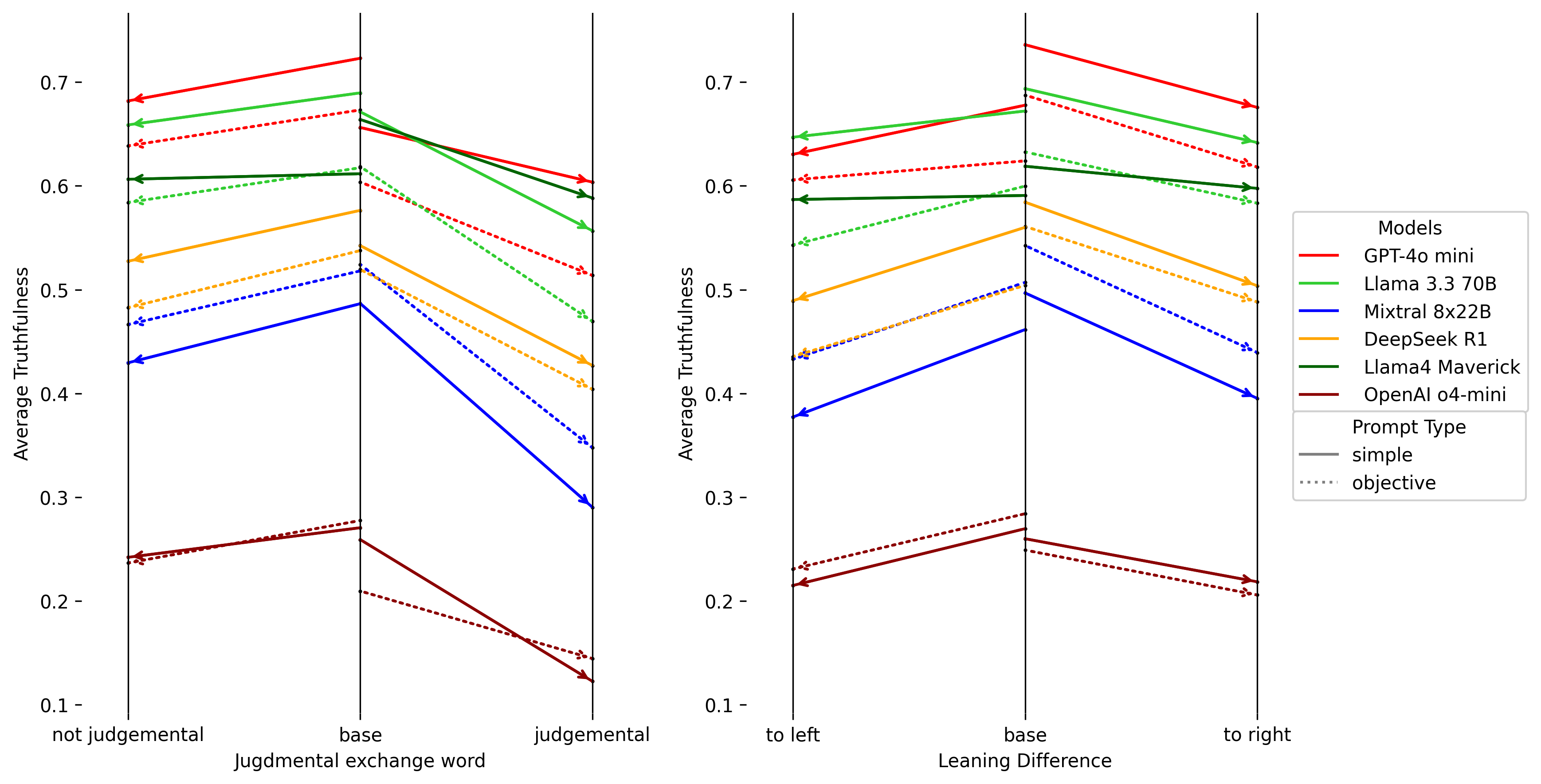}
  \caption{The impact of judgmental shifted words and political leaning on predicted truthfulness for different models and prompts.
  The solid arrows refer to the setting with a simple prompt and the dashed lines refer to the objective prompt where objectivity is emphasized.
  The left subplot shows the impact of judgmental shifted words.
  We can see how the average truthfulness changes when replacing the base word with non-judgmental word and a judgmental word.
  The truthfulness decreases significantly more for almost all models when the replacement word is judgmental (see Table~\ref{tab:significance_table_small}).
  The right subplot shows the impact of political leaning.
  We observe how the average truthfulness changes when replacing the base word with a more left-leaning word and a more right-leaning word.
  A decrease of predicted truthfulness can be detected in both directions, whereas a stark difference between left and right is not apparent.
  Overall, the impact of judgmental words on average truthfulness is bigger than political leaning.}
  \label{fig:Arrow_combined}
\end{figure*}

Table~\ref{tab:significance_table_small} shows for which model and prompt combinations judgmental words or political leaning are significant according to our logistic regression.
Table~\ref{tab:significance_table} in Appendix \ref{sec:appendix_results} presents detailed results across all experimental settings. Llama4 Maverick failed to generate valid outputs for 15\% of the simple prompts and 92\% of the objective prompts. Due to the limited data available for the Llama4 Maverick objective prompts, these outputs were excluded from the analysis. OpenAI o4-mini failed to generate valid output for 21.2\% of simple prompts and 24.5\% of objective prompts. The rest of the models refused to respond to both prompts less than 5\% of the time.
For all models and both prompts, we observe a significant effect of judgmental language, clearly indicating that claims containing judgmental words are more likely to be perceived as false. This effect is non-significant only once at the level of individual axes.
However, the political leaning shift has only a limited significant impact. 

For OpenAI o4-mini, when prompted with the simple prompt, the p-value of 0.0035 on the social axes is significant and when prompted with the objective prompt the p-value 0.0035 on the economic axes is significant. The coefficients for both prompts are positive for the economic axis and negative for the social axis. This means that OpenAI o4-mini estimated left-leaning claims as more truthful in social topics and right-leaning claims in economic topics.

The analysis for Llama 3.3 70B shows for the simple prompt a significance when considering all claims ($\text{p} = 0.0044$, $\text{coeff} = -0.2282$) and for the objective prompt when considering economic claims ($\text{p} = 0.0162$, $\text{coeff} = 0.2901$). In this case, the sign of the coefficients flips from being negative for the simple prompt in the combined dimension and being positive for the objective prompt in the economic dimensions. This means that calling for objectivity in the prompt changed the model to favour right-leaning claims on the economic axis instead of left-leaning claims on the combined axis.

The third model with significant political shift results is Llama4 Maverick. Evaluating only the simple prompt, the social axis, and the combined testing resulted in p-values of 0.0036 and 0.0002, with negative (left-leaning) coefficients.

Figure~\ref{fig:Arrow_combined} shows the deviations in the average of truthfulness ratings for each experimental setting, regarding political leaning and presence of judgment. 
OpenAI’s GPT-4o mini model provides the most optimistic evaluations, with a mean truthfulness score of 0.76 in the baseline condition. In contrast, OpenAI o4-mini stands out as the most pessimistic, with an average score of 0.27. 
In case of a leaning difference, for all LLMs except OpenAI o4-mini, we observe that base claims of left-shifted minimal pairs are, on average, rated as less truthful than the base claims of the right-shifted minimal pairs. In case of the presence of judgement, we observe that mostly the truthfulness of base claims for minimal pairs without judgement is higher compared to minimal pairs with judgement.

\begin{figure}[ht]
  \includegraphics[width=\columnwidth]{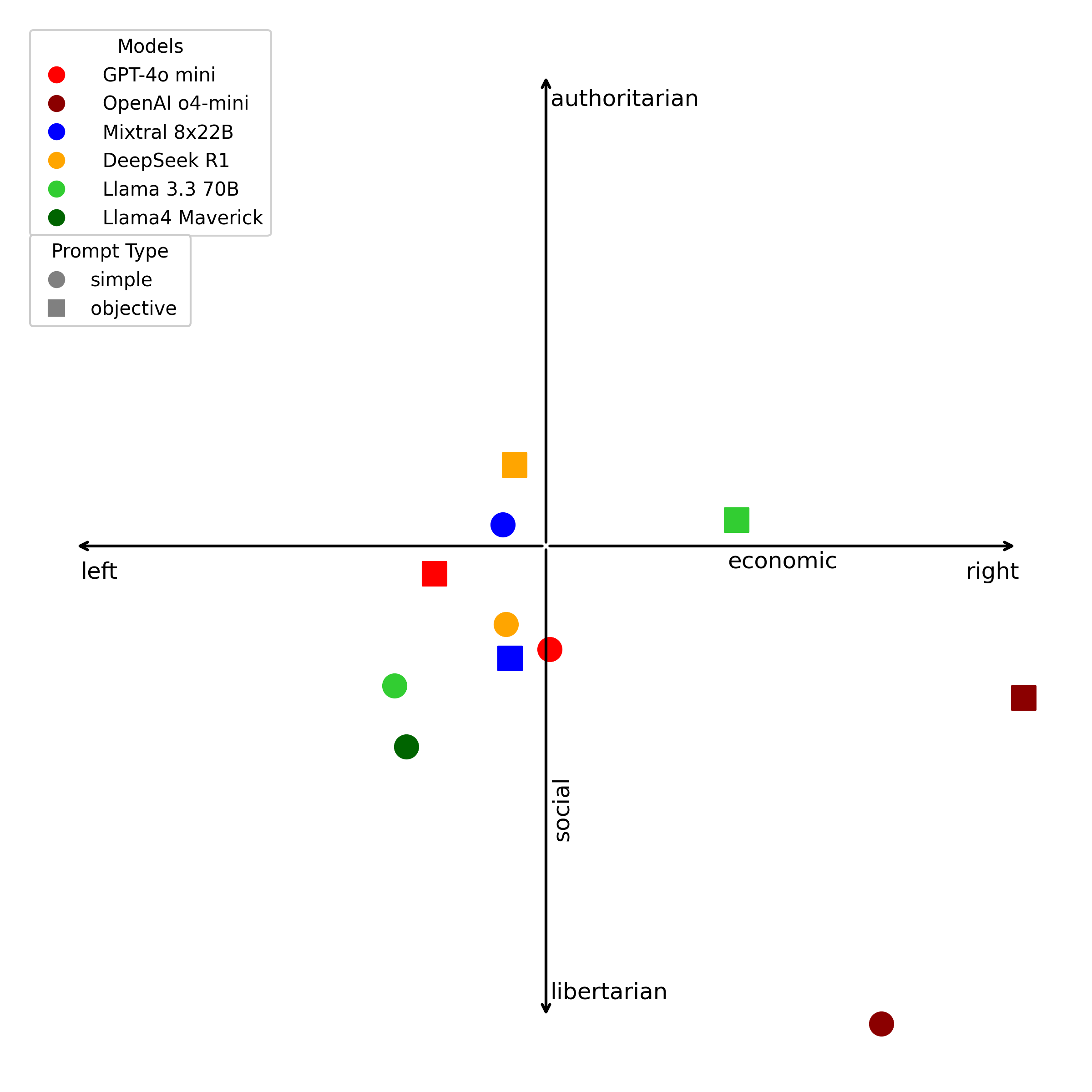}
  \caption{Aggregated results of all models along both political dimensions.
  The scaling is arbitrary, as our logistic model can yield any real number coefficient. Most models show left-libertarian tendencies, with Llama4 Maverick being the clearest example, while Llama 3.3 70B in the objective prompt setting and OpenAI o4-mini in both prompt settings show tendencies towards the right on the economic scale. We observe varying effects with regard to the objective prompt.}
  \label{fig:political_compass}
\end{figure}

In Figure~\ref{fig:political_compass}, we plot the coefficients for both axes in a two-dimensional scatter plot. The data does not reveal a consistent pattern of objective prompts producing more centered estimations relative to simple prompts.

In response to Research Question 1, our findings indicate that political connotation only partially influences LLMs’ perception of truth and that this effect varies depending on the prompt used.
However, we consistently observe a significant negative impact of judgmental language.

For Research Question 2, no big difference in impact of political leaning can be detected, as this signal was only present in few cases. 
Furthermore, the impact of judgmental words remain present in the results of the objective prompt, indicating that explicitly prompting for objectivism does not change the outcome.

\section{Discussion}

In this study, we present a novel German-language dataset that combines fact-checking tasks with political bias detection.
Given the complexity of the topic, a variety of approaches are conceivable.
For our methodology, we selected a minimalistic strategy: replacing targeted words in short, politically charged statements.
This method provides an initial, focused perspective on how political language may influence the truthfulness assessments of LLMs.
However, these results should not be generalized beyond the scope of this dataset.
Instead, they serve as a preliminary indication of potential political tendencies in LLM behavior under specific prompt and phrasing conditions.

One of the most unexpected findings of our study is the significant role judgmental words plays in how LLMs assess truthfulness. 
Words with judgmental undertones appear to skew the models' truth evaluations more than the actual political leaning of a statement.
This may be because emotionally loaded language is less frequent in factually reliable texts.
The political leaning of a statement had no consistent significant effect on the binary classification of a claim as ``True'' or ``False''.
Three of six models showed significant tendencies varying in topical dimension and leaning direction.

Shifted claims resulted in lower average truthfulness scores across all tested models. This could be explained by three factors. First, the base claims might be present in the training data of the models. Second, in general politically connotated words are more likely to be associated with misinformation independent of their political leaning. Third, when changing a word in a true sentence it is more likely to become false compared to a false claim becoming true. 

The result from the Llama4 Maverick can be interpreted optimistically. While Llama 3.3 70B when paired with the objective prompt, didn't reduce left-leaning political bias but changed towards the opposite direction, Llama4 Maverick instead refused to answer because of insufficient context. Such refusals may reflect a more cautious and context-aware reasoning process.

We caution against over-interpreting the results from our experiment.
Our work does not show that LLMs do not contain political biases, whether from specific models or in general. 
We only show that political shift is not the major factor in the experiments on our dataset and offer an alternative explanation for some experiments where political bias is detected.

\section{Conclusion}

This study provides an initial exploration of how LLMs evaluate politically charged statements in German.
We found that judgmental wording has a stronger impact on truthfulness predictions than political leaning. 
For three of the six models, we observed that their ratings of statements varied depending on the topic and the ideological leaning of the shifted words. OpenAI's o4-mini tends to favour economically right and socially libertarian viewpoints, Llama4 Maverick exhibit a left-leaning bias, and Llama 3.3 70B  shows varying bias directions based on the prompt.

While Llama4 Maverick shows a promising step toward mitigating bias, our results also highlight the limitations of current models in reliably separating political tone from factual content. As such, this dataset should not be seen as a benchmark, but as a starting point for deeper investigations into LLM behaviour.

Future work should focus on broader datasets and more diverse prompts to better understand how political and emotional cues influence truthfulness assessments.
Building on this work, it would be interesting to research the extent to which a model's perception of sentiment influences its political bias.

\section*{Limitations}

While our study provides valuable insights, several limitations should be considered.
We are aware that it is a difficult task to find X-phemisms that do not change the meaning of the sentence in any way.
Our dataset consists exclusively of German-language texts, which may limit the generalizability of our findings to other languages and cultural contexts.
We observe a rather low inter-annotator agreement, suggesting that annotating a political shift is difficult and subjective.
Thus, the quality of samples, both for statement similarity and political shift may be comparatively low.

Another limitation stems from the use of LLMs to generate word pairs.
Given that LLMs are known to exhibit a left-leaning bias, it is possible that euphemisms were more frequently chosen from left-leaning discourse, while dysphemisms were drawn from right-leaning language.
This could influence the observed bias patterns and warrants further investigation in future studies.
Furthermore, the annotations may be influenced by the annotators' clear left-leaning bias.

\section*{Acknowledgments}
We would like to thank Sebastian Möller, Leonhard Hennig and the anonymous reviewers for their feedback on the paper. 
We also thank our annotators Max Upravitelev and Maximilian Warsinke.
This work was supported by the German Ministry of Research, Technology and Space (BMFTR) as part of the projects newspolygraph (03RU2U151C) and TRAILS (01IW24005).

\bibliography{custom}

\begin{thebibliography}{34}
\providecommand{\natexlab}[1]{#1}

\bibitem[{Aksenov et~al.(2021)Aksenov, Bourgonje, Zaczynska, Ostendorff, Moreno-Schneider, and Rehm}]{aksenov_fine-grained_2021}
Dmitrii Aksenov, Peter Bourgonje, Karolina Zaczynska, Malte Ostendorff, Julian Moreno-Schneider, and Georg Rehm. 2021.
\newblock \href {https://doi.org/10.18653/v1/2021.woah-1.13} {Fine-grained {Classification} of {Political} {Bias} in {German} {News}: {A} {Data} {Set} and {Initial} {Experiments}}.
\newblock In \emph{Proceedings of the 5th {Workshop} on {Online} {Abuse} and {Harms} ({WOAH} 2021)}, pages 121--131, Online. Association for Computational Linguistics.

\bibitem[{Ashraf et~al.(2024)Ashraf, Bezzaoui, Andone, Markowetz, Fegert, and Flek}]{ashraf-2024-DefaktS:}
Shaina Ashraf, Isabel Bezzaoui, Ionut Andone, Alexander Markowetz, Jonas Fegert, and Lucie Flek. 2024.
\newblock \href {https://aclanthology.org/2024.lrec-main.409/} {Defakts: A fine-grained dataset for analyzing disinformation in german media}.
\newblock In \emph{Proceedings of The 2024 Joint International Conference, on Computational Linguistics, Language Resources and Evaluation}, Torino, Italia. European Language Resources Association.

\bibitem[{Baly et~al.(2019)Baly, Karadzhov, Saleh, Glass, and Nakov}]{balyMultiTaskOrdinalRegression2019a}
Ramy Baly, Georgi Karadzhov, Abdelrhman Saleh, James Glass, and Preslav Nakov. 2019.
\newblock \href {https://doi.org/10.18653/v1/N19-1216} {Multi-{Task} {Ordinal} {Regression} for {Jointly} {Predicting} the {Trustworthiness} and the {Leading} {Political} {Ideology} of {News} {Media}}.
\newblock In \emph{Proceedings of the 2019 {Conference} of the {North}}, pages 2109--2116, Minneapolis, Minnesota. Association for Computational Linguistics.

\bibitem[{Bleick et~al.(2024)Bleick, Feldhus, Burchardt, and M{\"o}ller}]{bleick2024german}
Maximilian Bleick, Nils Feldhus, Aljoscha Burchardt, and Sebastian M{\"o}ller. 2024.
\newblock \href {https://doi.org/10.18653/v1/2024.inlg-main.13} {{G}erman voter personas can radicalize {LLM} chatbots via the echo chamber effect}.
\newblock In \emph{Proceedings of the 17th International Natural Language Generation Conference}, pages 153--164, Tokyo, Japan. Association for Computational Linguistics.

\bibitem[{Brown et~al.(2020)Brown, Mann, Ryder, Subbiah, Kaplan, Dhariwal, Neelakantan, Shyam, Sastry, Askell, Agarwal, Herbert-Voss, Krueger, Henighan, Child, Ramesh, Ziegler, Wu, Winter, Hesse, Chen, Sigler, Litwin, Gray, Chess, Clark, Berner, McCandlish, Radford, Sutskever, and Amodei}]{brown2020language}
Tom Brown, Benjamin Mann, Nick Ryder, Melanie Subbiah, Jared~D Kaplan, Prafulla Dhariwal, Arvind Neelakantan, Pranav Shyam, Girish Sastry, Amanda Askell, Sandhini Agarwal, Ariel Herbert-Voss, Gretchen Krueger, Tom Henighan, Rewon Child, Aditya Ramesh, Daniel Ziegler, Jeffrey Wu, Clemens Winter, and 12 others. 2020.
\newblock \href {https://proceedings.neurips.cc/paper_files/paper/2020/file/1457c0d6bfcb4967418bfb8ac142f64a-Paper.pdf} {Language models are few-shot learners}.
\newblock In \emph{Advances in Neural Information Processing Systems}, volume~33, pages 1877--1901. Curran Associates, Inc.

\bibitem[{Bu{\'c}o(2022)}]{buco2022politische}
Veronika Bu{\'c}o. 2022.
\newblock \href {https://zir.nsk.hr/islandora/object/ffos:6138} {\emph{Politische Euphemismen in der deutschen Pressesprache}}.
\newblock Ph.D. thesis, Josip Juraj Strossmayer University of Osijek. Faculty of Humanities and Social Sciences. Department of German Language and Literature.

\bibitem[{Burkhardt(2010)}]{burkhardt2010euphemism}
Armin Burkhardt. 2010.
\newblock \href {https://doi.org/doi:10.1515/9783110230215.355} {\emph{Euphemism and truth}}, pages 355--372.
\newblock Walter de Gruyter Berlin, DEU.

\bibitem[{Chong and Druckman(2007)}]{chongFramingTheory2007}
Dennis Chong and James~N. Druckman. 2007.
\newblock \href {https://doi.org/10.1146/annurev.polisci.10.072805.103054} {Framing {Theory}}.
\newblock \emph{Annual Review of Political Science}, 10(1):103--126.

\bibitem[{Coeckelbergh(2025)}]{coeckelberghLLMsTruthDemocracy2025}
Mark Coeckelbergh. 2025.
\newblock \href {https://doi.org/10.1007/s11948-025-00529-0} {{LLMs}, {Truth}, and {Democracy}: {An} {Overview} of {Risks}}.
\newblock \emph{Science and Engineering Ethics}, 31(1):4.

\bibitem[{Danks and London(2017)}]{danksAlgorithmicBiasAutonomous2017}
David Danks and Alex~John London. 2017.
\newblock \href {https://doi.org/10.24963/ijcai.2017/654} {Algorithmic {Bias} in {Autonomous} {Systems}}.
\newblock In \emph{Proceedings of the {Twenty}-{Sixth} {International} {Joint} {Conference} on {Artificial} {Intelligence}}, pages 4691--4697, Melbourne, Australia. International Joint Conferences on Artificial Intelligence Organization.

\bibitem[{Dev et~al.(2020)Dev, Li, Phillips, and Srikumar}]{dev2020measuring}
Sunipa Dev, Tao Li, Jeff~M. Phillips, and Vivek Srikumar. 2020.
\newblock \href {https://ojs.aaai.org/index.php/AAAI/article/view/6267} {On measuring and mitigating biased inferences of word embeddings}.
\newblock \emph{Proceedings of the AAAI Conference on Artificial Intelligence}, 34(05):7659--7666.

\bibitem[{Durmus et~al.(2023)Durmus, Nguyen, Liao, Schiefer, Askell, Bakhtin, Chen, Hatfield-Dodds, Hernandez, Joseph, Lovitt, McCandlish, Sikder, Tamkin, Thamkul, Kaplan, Clark, and Ganguli}]{durmusMeasuringRepresentationSubjective2023}
Esin Durmus, Karina Nguyen, Thomas~I. Liao, Nicholas Schiefer, Amanda Askell, Anton Bakhtin, Carol Chen, Zac Hatfield-Dodds, Danny Hernandez, Nicholas Joseph, Liane Lovitt, Sam McCandlish, Orowa Sikder, Alex Tamkin, Janel Thamkul, Jared Kaplan, Jack Clark, and Deep Ganguli. 2023.
\newblock \href {https://doi.org/10.48550/ARXIV.2306.16388} {Towards {Measuring} the {Representation} of {Subjective} {Global} {Opinions} in {Language} {Models}}.
\newblock \emph{arXiv preprint}.
\newblock Version Number: 2.

\bibitem[{French et~al.(2023)French, Storey, and Wallace}]{frenchImpactCognitiveBiases2023}
Aaron~M. French, Veda~C. Storey, and Linda Wallace. 2023.
\newblock \href {https://doi.org/10.1080/0960085X.2023.2272608} {The impact of cognitive biases on the believability of fake news}.
\newblock \emph{European Journal of Information Systems}, pages 1--22.

\bibitem[{Gallegos et~al.(2024)Gallegos, Rossi, Barrow, Tanjim, Kim, Dernoncourt, Yu, Zhang, and Ahmed}]{gallegos2024bias}
Isabel~O Gallegos, Ryan~A Rossi, Joe Barrow, Md~Mehrab Tanjim, Sungchul Kim, Franck Dernoncourt, Tong Yu, Ruiyi Zhang, and Nesreen~K Ahmed. 2024.
\newblock \href
  {https://watermark02.silverchair.com/coli_a_00524.pdf?token=AQECAHi208BE49Ooan9kkhW_Ercy7Dm3ZL_9Cf3qfKAc485ysgAAAz0wggM5BgkqhkiG9w0BBwagggMqMIIDJgIBADCCAx8GCSqGSIb3DQEHATAeBglghkgBZQMEAS4wEQQMuyGHGmzvS9itdFDtAgEQgIIC8B7f5wkPddJ1GmHAYVfCZEBAY0ECEmREfj6x_Zn8a9GzPhf94UC2bsHIb58dvkKiQE4_-drtDvP8D5FuHeb5ygZWJP9QPA_75L_miwbXY4UQE9tiz-GnHYn8koPl3wkom0MlIyGv3mRdRrX0e8_I3YGVAq8owuy1Bl36LbSQ2Dr4l7SgSNLjmAzCCv5nQe3c3b_HAJLsCXg2Jbix-oDr3NEBm25CkmIxE2OzUq3Nbjalj9Gyt-8VDdei6p6BRJvFWsQXjGJPPtoee5UHWXUW9R4Bme9LhR2rzaSZzf1pe2q9EbJotWRbygtAQ7Yndh-z8QtSoyXlqiiAbwHOEwaSulokmqNs09WESPhp1CIxO7jmNdR0GxQHVp_Sjhr6tmjFS23IhfgQ54zs2qnCGr-LMEbGJ1DumP5-X05Jra6yvBObEZlQkbcW0BHF27sS4DxWaTRGyIZCO1QKmKQmFK1ciMR0aYQM9K1b8yftVKrP1zt_zaDSJbBWhRrQVN-iXLQv05PhPs9WIHwpFNaSx6VRDpGGydmwV-vBImX65rMQnwhlUbq2ajHGrggL5M0Cw_LGcdms2DuLVLtL7xEOH6RbhbEZbHw-LNoUTrL3vkEWJT52pwOjUlK7NHmu6QOCRJoE8r0x01f0lH837GmcAaeGClMdzCtIqWY6d148OHmOnv2sDeHmH0PM74_92qS14cDujkyWdiPaq1uHXzc6kCkwmZP61qjKaBhsf-7QP-eAni--qnkS_OcokeUayGwnPhu1TbFoJ2Nxend4OIbk2mo_91rGgFNP2xdl5AVCjPg5vwFkkvM1ei3JjpK7kgfzW1gymHEaTRdaiAiNxl402mWuMPidEJGzYwkTuQRqh39x9kIgJ6cq6w-JJV1YkT1n55WSTbYNf-OQbYt79dMGbXpHFjloBGHHYU6Puir2LFRs93aBYU4kx5QgXIgDoqadcX6tIu48yzn1fn6wj7QbcFOOE-TNcp5oeMVYvj7Q2vcBP322}
  {Bias and fairness in large language models: A survey}.
\newblock \emph{Computational Linguistics}, 50(3):1097--1179.

\bibitem[{Guo et~al.(2025)Guo, Yang, Zhang, Song, Zhang, Xu, Zhu, Ma, Wang, Bi et~al.}]{guo2025deepseek}
Daya Guo, Dejian Yang, Haowei Zhang, Junxiao Song, Ruoyu Zhang, Runxin Xu, Qihao Zhu, Shirong Ma, Peiyi Wang, Xiao Bi, and 1 others. 2025.
\newblock \href {https://arxiv.org/abs/2501.12948} {Deepseek-r1: Incentivizing reasoning capability in llms via reinforcement learning}.
\newblock \emph{arXiv preprint arXiv:2501.12948}.

\bibitem[{Hall(2001)}]{hall2001foucault}
Stuart Hall. 2001.
\newblock Foucault: Power, knowledge and discourse.
\newblock \emph{Discourse theory and practice: A reader}, 72.

\bibitem[{Hartmann et~al.(2023)Hartmann, Schwenzow, and Witte}]{hartmannPoliticalIdeologyConversational2023}
Jochen Hartmann, Jasper Schwenzow, and Maximilian Witte. 2023.
\newblock \href {https://doi.org/10.48550/ARXIV.2301.01768} {The political ideology of conversational {AI}: {Converging} evidence on {ChatGPT}'s pro-environmental, left-libertarian orientation}.
\newblock \emph{arXiv preprint}.
\newblock Version Number: 1.

\bibitem[{Havryliv(2021)}]{havryliv2021sprache}
Oksana Havryliv. 2021.
\newblock \href {https://www.ceeol.com/search/article-detail?id=1016295} {Sprache und corona}.
\newblock \emph{Linguistische Treffen in Wroc{\l}aw}, 20(2):69--88.

\bibitem[{Huang et~al.(2020)Huang, Zhang, Jiang, Stanforth, Welbl, Rae, Maini, Yogatama, and Kohli}]{huang2019reducing}
Po-Sen Huang, Huan Zhang, Ray Jiang, Robert Stanforth, Johannes Welbl, Jack Rae, Vishal Maini, Dani Yogatama, and Pushmeet Kohli. 2020.
\newblock \href {https://doi.org/10.18653/v1/2020.findings-emnlp.7} {Reducing sentiment bias in language models via counterfactual evaluation}.
\newblock In \emph{Findings of the Association for Computational Linguistics: EMNLP 2020}, pages 65--83, Online. Association for Computational Linguistics.

\bibitem[{Jiang et~al.(2024)Jiang, Sablayrolles, Roux, Mensch, Savary, Bamford, Chaplot, Casas, Hanna, Bressand et~al.}]{jiang2024mixtral}
Albert~Q Jiang, Alexandre Sablayrolles, Antoine Roux, Arthur Mensch, Blanche Savary, Chris Bamford, Devendra~Singh Chaplot, Diego de~las Casas, Emma~Bou Hanna, Florian Bressand, and 1 others. 2024.
\newblock \href {https://doi.org/10.48550/arXiv.2401.04088} {Mixtral of experts}.
\newblock \emph{arXiv preprint arXiv:2401.04088}.

\bibitem[{Liu et~al.(2025)Liu, Panwang, and Gu}]{liu2025turning}
Yifei Liu, Yuang Panwang, and Chao Gu. 2025.
\newblock \href {https://doi.org/10.1057/s41599-025-04465-z} {“{T}urning right”? an experimental study on the political value shift in large language models}.
\newblock \emph{Humanities and Social Sciences Communications}, 12(1):1--10.

\bibitem[{McGee(2023)}]{mcgeeChatGptBiased2023}
Robert~W. McGee. 2023.
\newblock \href {https://doi.org/10.2139/ssrn.4359405} {Is {Chat} {Gpt} {Biased} {Against} {Conservatives}? {An} {Empirical} {Study}}.
\newblock \emph{SSRN Electronic Journal}.

\bibitem[{Mohtaj et~al.(2024)Mohtaj, Nizamoglu, Sahitaj, Schmitt, Jakob, and M\"{o}ller}]{newspolyml}
Salar Mohtaj, Ata Nizamoglu, Premtim Sahitaj, Vera Schmitt, Charlott Jakob, and Sebastian M\"{o}ller. 2024.
\newblock \href {https://doi.org/10.1145/3643491.3660290} {Newspolyml: Multi-lingual european news fake assessment dataset}.
\newblock In \emph{Proceedings of the 3rd ACM International Workshop on Multimedia AI against Disinformation}, MAD '24, page 82–90, New York, NY, USA. Association for Computing Machinery.

\bibitem[{Motoki et~al.(2023)Motoki, Pinho~Neto, and Rodrigues}]{motokiMoreHumanHuman2023}
Fabio Motoki, Valdemar Pinho~Neto, and Victor Rodrigues. 2023.
\newblock \href {https://doi.org/10.1007/s11127-023-01097-2} {More human than human: measuring {ChatGPT} political bias}.
\newblock \emph{Public Choice}.

\bibitem[{Park et~al.(2018)Park, Shin, and Fung}]{park2018reducing}
Ji~Ho Park, Jamin Shin, and Pascale Fung. 2018.
\newblock \href {https://doi.org/10.18653/v1/D18-1302} {Reducing gender bias in abusive language detection}.
\newblock In \emph{Proceedings of the 2018 Conference on Empirical Methods in Natural Language Processing}, pages 2799--2804, Brussels, Belgium. Association for Computational Linguistics.

\bibitem[{Risch et~al.(2021)Risch, Stoll, Wilms, and Wiegand}]{risch_overview_2021}
Julian Risch, Anke Stoll, Lena Wilms, and Michael Wiegand. 2021.
\newblock \href {https://aclanthology.org/2021.germeval-1.1} {Overview of the {GermEval} 2021 {Shared} {Task} on the {Identification} of {Toxic}, {Engaging}, and {Fact}-{Claiming} {Comments}}.
\newblock In \emph{Proceedings of the {GermEval} 2021 {Shared} {Task} on the {Identification} of {Toxic}, {Engaging}, and {Fact}-{Claiming} {Comments}}, Duesseldorf, Germany. Association for Computational Linguistics.

\bibitem[{Rozado(2023)}]{rozadoPoliticalBiasesChatGPT2023}
David Rozado. 2023.
\newblock \href {https://doi.org/10.3390/socsci12030148} {The {Political} {Biases} of {ChatGPT}}.
\newblock \emph{Social Sciences}, 12(3):148.

\bibitem[{Rozado(2024)}]{rozadoPoliticalPreferencesLLMs2024}
David Rozado. 2024.
\newblock \href {https://doi.org/10.1371/journal.pone.0306621} {The {Political} {Preferences} of {LLMs}}.
\newblock \emph{PLoS ONE 19(7): e0306621}.
\newblock Version Number: 2.

\bibitem[{Rutinowski et~al.(2024)Rutinowski, Franke, Endendyk, Dormuth, Roidl, and Pauly}]{utinowskiSelfPerceptionPoliticalBiases2024}
Jérôme Rutinowski, Sven Franke, Jan Endendyk, Ina Dormuth, Moritz Roidl, and Markus Pauly. 2024.
\newblock \href {https://doi.org/10.1155/2024/7115633} {The {Self}-{Perception} and {Political} {Biases} of {ChatGPT}}.
\newblock \emph{Human Behavior and Emerging Technologies}, 2024:1--9.

\bibitem[{Röttger et~al.(2024)Röttger, Hofmann, Pyatkin, Hinck, Kirk, Schuetze, and Hovy}]{rottgerPoliticalCompassSpinning2024}
Paul Röttger, Valentin Hofmann, Valentina Pyatkin, Musashi Hinck, Hannah Kirk, Hinrich Schuetze, and Dirk Hovy. 2024.
\newblock \href {https://doi.org/10.18653/v1/2024.acl-long.816} {Political {Compass} or {Spinning} {Arrow}? {Towards} {More} {Meaningful} {Evaluations} for {Values} and {Opinions} in {Large} {Language} {Models}}.
\newblock In \emph{Proceedings of the 62nd {Annual} {Meeting} of the {Association} for {Computational} {Linguistics} ({Volume} 1: {Long} {Papers})}, pages 15295--15311, Bangkok, Thailand. Association for Computational Linguistics.

\bibitem[{Shen and Rose(2021)}]{shenWhatSoundsRight2021}
Qinlan Shen and Carolyn Rose. 2021.
\newblock \href {https://doi.org/10.18653/v1/2021.eacl-main.152} {What {Sounds} “{Right}” to {Me}? {Experiential} {Factors} in the {Perception} of {Political} {Ideology}}.
\newblock In \emph{Proceedings of the 16th {Conference} of the {European} {Chapter} of the {Association} for {Computational} {Linguistics}: {Main} {Volume}}, pages 1762--1771, Online. Association for Computational Linguistics.

\bibitem[{Touvron et~al.(2023)Touvron, Lavril, Izacard, Martinet, Lachaux, Lacroix, Rozi{\`e}re, Goyal, Hambro, Azhar et~al.}]{touvron2023llama}
Hugo Touvron, Thibaut Lavril, Gautier Izacard, Xavier Martinet, Marie-Anne Lachaux, Timoth{\'e}e Lacroix, Baptiste Rozi{\`e}re, Naman Goyal, Eric Hambro, Faisal Azhar, and 1 others. 2023.
\newblock \href {https://doi.org/10.48550/arXiv.2302.13971} {Llama: Open and efficient foundation language models}.
\newblock \emph{arXiv preprint arXiv:2302.13971}.

\bibitem[{Urman and Makhortykh(2025)}]{urmanSilenceLLMsCrosslingual2025}
Aleksandra Urman and Mykola Makhortykh. 2025.
\newblock \href {https://doi.org/10.1016/j.tele.2024.102211} {The silence of the {LLMs}: {Cross}-lingual analysis of guardrail-related political bias and false information prevalence in {ChatGPT}, {Google} {Bard} ({Gemini}), and {Bing} {Chat}}.
\newblock \emph{Telematics and Informatics}, 96:102211.

\bibitem[{Van Der~Linden et~al.(2020)Van Der~Linden, Panagopoulos, and Roozenbeek}]{vanderlindenYouAreFake2020}
Sander Van Der~Linden, Costas Panagopoulos, and Jon Roozenbeek. 2020.
\newblock \href {https://doi.org/10.1177/0163443720906992} {You are fake news: political bias in perceptions of fake news}.
\newblock \emph{Media, Culture \& Society}, 42(3):460--470.

\end{thebibliography}
.

\clearpage
\onecolumn

\appendix

\section{Results}
\label{sec:appendix_results}

\small
\begin{table}[H]
    \centering

    \small
    \begin{tabular}{l|l|l|cccccccc}
        \toprule
         \textbf{LLM} & \textbf{Prompt} & \textbf{Dimension} & \multicolumn{2}{c}{\textbf{Axis}} & \multicolumn{2}{c}{\textbf{Jugment base}} & \multicolumn{2}{c}{\textbf{Judgement exchange}} & \multicolumn{2}{c}{\textbf{Leaning shift}}\\
        \midrule
        & & & \textbf{coef} & \textbf{p} & \textbf{coef} & \textbf{p} & \textbf{coef} & \textbf{p} & \textbf{coef} & \textbf{p}  \\
\cmidrule{1-11}
 \multirow{6}{*}{GPT-4o mini} & \multirow{3}{*}{simple} & both & \textbf{-0.9260} & 0.0000 & -0.7294 & 0.1139 & \textbf{-0.8725} & 0.0000 & -0.0958 & 0.1874 \\
  &  & social & - & - & -1.2005 & 0.1487 & \textbf{-0.8954} & 0.0001 & -0.1569 & 0.0971 \\
  &  & economic & - & - & -0.6626 & 0.2563 & \textbf{-0.8789} & 0.0001 & 0.0053 & 0.9640 \\
\cmidrule{2-11}
  & \multirow{3}{*}{advanced} & both & \textbf{-0.6691} & 0.0000 & 0.3480 & 0.4518 & \textbf{-1.0740} & 0.0000 & -0.1261 & 0.0703 \\
  &  & social & - & - & 0.3875 & 0.5854 & \textbf{-1.1686} & 0.0000 & -0.0421 & 0.6670 \\
  &  & economic & - & - & 0.5152 & 0.4264 & \textbf{-1.0041} & 0.0000 & -0.1694 & 0.0993 \\
\cmidrule{1-11}
 \multirow{6}{*}{OpenAI o4-mini} & \multirow{3}{*}{simple} & both & 0.0371 & 0.8901 & \textbf{2.2086} & 0.0015 & \textbf{-2.0564} & 0.0000 & -0.0408 & 0.7907 \\
  &  & social & - & - & 33.9025 & 1.0000 & \textbf{-1.7428} & 0.0006 & \textbf{-0.7274} & 0.0035 \\
  &  & economic & - & - & \textbf{2.7727} & 0.0011 & \textbf{-2.5152} & 0.0000 & 0.5104 & 0.0347 \\
\cmidrule{2-11}
  & \multirow{3}{*}{advanced} & both & \textbf{0.7851} & 0.0092 & 0.9859 & 0.1967 & \textbf{-1.2688} & 0.0008 & 0.2149 & 0.1834 \\
  &  & social & - & - & 1.2561 & 0.1768 & -0.9554 & 0.1205 & -0.2312 & 0.3351 \\
  &  & economic & - & - & -0.6567 & 0.5816 & \textbf{-1.2662} & 0.0120 & \textbf{0.7271} & 0.0035 \\
\cmidrule{1-11}
 \multirow{6}{*}{Mixtral 8x22B} & \multirow{3}{*}{simple} & both & \textbf{-0.4838} & 0.0000 & 0.6461 & 0.1598 & \textbf{-1.3858} & 0.0000 & -0.0189 & 0.7763 \\
  &  & social & - & - & 0.5386 & 0.5230 & \textbf{-1.3974} & 0.0000 & 0.0326 & 0.7322 \\
  &  & economic & - & - & 0.7882 & 0.1694 & \textbf{-1.3575} & 0.0000 & -0.0658 & 0.4859 \\
\cmidrule{2-11}
  & \multirow{3}{*}{advanced} & both & \textbf{-0.3998} & 0.0005 & \textbf{1.1739} & 0.0151 & \textbf{-1.2340} & 0.0000 & -0.1097 & 0.0960 \\
  &  & social & - & - & 0.8488 & 0.3513 & \textbf{-1.3661} & 0.0000 & -0.1710 & 0.0771 \\
  &  & economic & - & - & 1.3590 & 0.0257 & \textbf{-1.1472} & 0.0000 & -0.0550 & 0.5494 \\
\cmidrule{1-11}
 \multirow{6}{*}{DeepSeek R1} & \multirow{3}{*}{simple} & both & \textbf{-0.4768} & 0.0000 & 0.4486 & 0.3113 & \textbf{-0.9226} & 0.0000 & -0.1001 & 0.1275 \\
  &  & social & - & - & 0.7349 & 0.2655 & \textbf{-1.1453} & 0.0000 & -0.1193 & 0.2109 \\
  &  & economic & - & - & 0.2404 & 0.6710 & \textbf{-0.7735} & 0.0000 & -0.0612 & 0.5088 \\
\cmidrule{2-11}
  & \multirow{3}{*}{advanced} & both & -0.1674 & 0.1906 & 0.5500 & 0.2833 & \textbf{-0.7867} & 0.0000 & 0.0221 & 0.7622 \\
  &  & social & - & - & 1.4434 & 0.0448 & \textbf{-0.6610} & 0.0161 & 0.1236 & 0.2456 \\
  &  & economic & - & - & 0.2095 & 0.7249 & \textbf{-0.8125} & 0.0001 & -0.0479 & 0.6406 \\
\cmidrule{1-11}
 \multirow{6}{*}{Llama 3.3 70B} & \multirow{3}{*}{simple} & both & \textbf{-0.7100} & 0.0000 & 0.4696 & 0.4374 & \textbf{-1.0307} & 0.0000 & \textbf{-0.2282} & 0.0044 \\
  &  & social & - & - & 0.4186 & 0.7391 & \textbf{-1.0642} & 0.0001 & -0.2126 & 0.0603 \\
  &  & economic & - & - & 0.6448 & 0.3789 & \textbf{-1.0110} & 0.0000 & -0.2307 & 0.0466 \\
\cmidrule{2-11}
  & \multirow{3}{*}{advanced} & both & \textbf{-0.3422} & 0.0176 & 0.8265 & 0.2027 & \textbf{-1.4186} & 0.0000 & 0.1489 & 0.0647 \\
  &  & social & - & - & 0.5177 & 0.6449 & \textbf{-1.5751} & 0.0000 & 0.0395 & 0.7214 \\
  &  & economic & - & - & 0.8835 & 0.2955 & \textbf{-1.3258} & 0.0000 & \textbf{0.2901} & 0.0162 \\
\cmidrule{1-11}
 \multirow{3}{*}{Llama4 Maverick} & \multirow{3}{*}{simple} & both & \textbf{-0.7138} & 0.0000 & 0.2324 & 0.6284 & \textbf{-0.6106} & 0.0002 & \textbf{-0.2794} & 0.0002 \\
  &  & social & - & - & 0.2285 & 0.7736 & \textbf{-0.6439} & 0.0171 & \textbf{-0.3056} & 0.0036 \\
  &  & economic & - & - & 0.2590 & 0.6830 & \textbf{-0.5603} & 0.0085 & -0.2124 & 0.0509 \\

        \bottomrule
    \end{tabular}
    \caption{Comparison of Policy Models and LLMs}
    \label{tab:significance_table}
\end{table}

\section{X-phemisms Sources}
\label{sec:appendix_sources}
\begin{table}[h!]
    \centering
    \label{tab:manual_euphemisms}
    \small
    \resizebox{\linewidth}{!}{%
    \begin{tabular}{l}
        \toprule
        \citet{buco2022politische} \\
        \citet{havryliv2021sprache} \\
        \url{https://bastiansick.de/kolumnen/zwiebelfisch/75-euphemismen-aus-alltagssprache-wirtschaft-und-politik/}\\
        \url{https://de.wikipedia.org/wiki/Liste_von_Euphemismen}\\
        \url{https://www.bpb.de/themen/parteien/rechtspopulismus/240831/rechtspopulistische-lexik-und-die-grenzen-des-sagbaren/} \\
        \url{https://www.pichler-training.at/dysphemismen-sprachliche-ausdruecke-mit-abwertungen/} \\
        \url{https://www.srf.ch/kultur/gesellschaft-religion/macht-der-sprache-was-wir-meinen-wenn-wir-fluechtling-sagen} \\
        \url{https://www.tagesspiegel.de/wissen/so-werden-die-grenzen-des-sagbaren-verschoben-4202638.html} \\
        \bottomrule
    \end{tabular}
    }
    \caption{We used these sources for X-phemisms. 
    See Section \ref{sec:word_pair_collection} for details on how these were used.}
\end{table}

\section{Prompts}
\label{sec:appendix_prompts}

\subsection{X-phemism Generation}
\textit{``Ich möchte X-phemismen, die in der deutschen Politik verwendet werden. Zähle 100 auf mit dem dazugehörigen neutralen Wort und sage, ob es Dysphemismen oder Euphemismen sind und welches politische Lager dieses Wort benutzt. Gib eine Tabelle zurück mit einem Paar aus neutralem Begriff und X-phemismus pro Zeile.''}\\

\noindent\textit{``I want X-phemisms that are used in German politics. List 100 with the corresponding neutral word and say whether they are dysphemisms or euphemisms and which political camp uses this word. Return a table with a pair of neutral terms and X-phemisms per line.''}

\subsection{Grammar Checking}
\textit{System prompt: ``Du bist ein hilfreicher Assistent und hilfst bei einer wissenschaftlichen Untersuchung zu Wörtern und Ausdrücken, die von verschiedenen politischen Lagern verwendet werden.''}\\
\textit{User prompt: ``Bitte korrigiere die Grammatik des folgenden Satzes und gib nur den korrigierten Satz zurück ohne Erklärungen oder Kommentare. Du sollst keine inhaltlichen Veränderungen oder Verbesserungen vornehmen, auch wenn der Satz inhaltlich keinen Sinn ergibt. Es sollen keine Wörter ausgetauscht oder hinzugefügt werden, nur Artikelanpassungen und Wortendungsanpassungen. Es kann sein das manche Begriffe politisch aufgeladen sind bitte nehme trotzdem keine inhaltlichen Korrekturen vor. Falls der Satz bereits korrekt ist gib den Satz unverändert zurück, ebenfalls ohne Erklärungen oder Kommentare.\{sentence\}''}\\

\noindent\textit{System prompt: ``You are a helpful assistant and are assisting with a scientific study on words and expressions used by different political camps.}\\
\textit{User prompt: ``Please correct the grammar of the following sentence and return only the corrected sentence without explanations or comments. You should not make any changes or improvements to the content, even if the sentence does not make sense. No words should be replaced or added, only articles and word endings should be adjusted. Some terms may be politically charged, but please do not make any corrections to the content. If the sentence is already correct, return it unchanged, again without explanations or comments.\{sentence\}''}

\section{Annotations}
\label{sec:appendix_annotations}

\subsection{Annotation Results}
\small
\begin{longtable}{llcccc}
\caption{Word Pair Shifts} \\
\toprule
First Word & Second Word &  Shift (->) & Axis & First Judg. & Second Judg. \\
\midrule
\endfirsthead

\toprule
First Word & Second Word &  Shift (->) & Axis & First Judg. & Second Judg. \\
\midrule
\endhead

\midrule
\endfoot

\bottomrule
\endlastfoot
\multicolumn{6}{c}{\textbf{Manual Collected Word Pairs}} \\
\midrule
Abgabe & Steuer & right & economic & 0 & 0 \\
Abriss & Rückbau & left & economic & 0 & 0 \\
Abschiebung & aufenthaltsbeendende Maßnahmen & authoritarian & social & 0 & 0 \\
Agrarbetrieb & Bauernhof & right & economic & 0 & 0 \\
Altersheim & Seniorenwohnheim & no shift & social & 0 & 0 \\
Angriff & Offensive & authoritarian & social & 0 & 0 \\
Angriffskrieg & militärische Spezialoperation & authoritarian & social & 0 & 0 \\
Atomenergie & Kernenergie & no shift & economic & 0 & 0 \\
Atomkraft & Kernenergie & no shift & economic & 0 & 0 \\
Ausgangssperre & Lockdown & no shift & social & 0 & 0 \\
Bande & Gruppe & libertarian & social & 0 & 0 \\
Berliner Mauer & Schandmauer & authoritarian & social & 0 & 1 \\
Berliner Mauer & antifaschistischer Schutzwall & libertarian & social & 0 & 1 \\
Bordell & Freudenhaus & authoritarian & social & 0 & 0 \\
Dame & Frau & libertarian & social & 0 & 0 \\
Deportation & Remigration & authoritarian & social & 0 & 0 \\
Drahtzieher & Verantwortlicher & libertarian & social & 0 & 0 \\
Endlager & Entsorungspark & right & economic & 0 & 0 \\
Entlassung & Umstrukturierung & right & economic & 0 & 0 \\
Finanzamt & stiller Teilhaber & right & economic & 0 & 0 \\
Flüchtling & Geflüchtete & libertarian & social & 0 & 0 \\
Flüchtlingsbewegung & Bevölkerungsaustausch & authoritarian & social & 0 & 0 \\
Flüchtlingsbewegung & Entdeutschung & authoritarian & social & 0 & 0 \\
Flüchtlingsbewegung & Flüchtlingsstrom & authoritarian & social & 0 & 0 \\
Folter & erweiterte Verhörmethoden & authoritarian & social & 0 & 0 \\
Gefallener & Kriegsopfer & libertarian & social & 0 & 0 \\
Geheimdienst & Nachrichtendienst & authoritarian & social & 0 & 0 \\
Gendern & Genderwahnsinn & authoritarian & social & 0 & 1 \\
Gesundheitskasse & Krankenkasse & authoritarian & social & 0 & 0 \\
Handelsvertreter & Außendienstmitarbeiter & left & economic & 0 & 0 \\
Hausmeister & Facility Manager & no shift & social & 0 & 0 \\
Häftling & Knacki & authoritarian & social & 0 & 1 \\
Konflikt & Krieg & libertarian & social & 0 & 0 \\
Konkurrent & Mitbewerber & libertarian & social & 0 & 0 \\
Konspiration & Verschwörung & no shift & social & 0 & 0 \\
Kopftuchträgerin & Kopftuchmädchen & authoritarian & social & 0 & 0 \\
Krieg & Verteidigungsfall & libertarian & social & 0 & 0 \\
Kriegsminister & Verteidigungsminister & libertarian & social & 0 & 0 \\
Lügenpresse & Medien & libertarian & social & 1 & 0 \\
Machthaber & Regierung & libertarian & social & 0 & 0 \\
Mittel & Waffen & libertarian & social & 0 & 0 \\
Norm & Regel & libertarian & social & 0 & 0 \\
Nullwachstum & Stagnation & no shift & economic & 0 & 0 \\
Obdachloser & Penner & authoritarian & social & 0 & 1 \\
Opposition & Widerstandskämpfer & no shift & social & 0 & 0 \\
Prostitution & käufliche Liebe & libertarian & social & 0 & 0 \\
Putzfrau & Raumpflegerin & libertarian & social & 0 & 0 \\
Rezession & Wachstumspause & right & economic & 0 & 0 \\
Rückgang & negative Zuwachsraten & no shift & economic & 0 & 0 \\
Sterben & ableben & no shift & social & 0 & 0 \\
Umsiedlung & Vertreibung & libertarian & social & 0 & 0 \\
Völkermord & ethnische Säuberung & authoritarian & social & 0 & 0 \\
Waldsterben & Absterben verschiedener Baumarten & right & economic & 0 & 0 \\
Werbung & Verbraucherinformationen & right & economic & 0 & 0 \\
abdanken & Aufgeben eines Amtes & no shift & none & 0 & 0 \\
arbeitslos & arbeitssuchend & left & economic & 0 & 0 \\
autoritär & durchsetzungsfähig & authoritarian & social & 0 & 0 \\
barrierefrei & behindertengerecht & authoritarian & social & 0 & 0 \\
bildungsfern & ungebildet & no shift & social & 0 & 0 \\
dickköpfig & beratungsresistent & no shift & social & 1 & 0 \\
hochpreisig & teuer & no shift & economic & 0 & 0 \\
in den Krieg ziehen & zu den Waffen greifen & authoritarian & social & 0 & 0 \\
in den besten Jahren & kurz vor dem Eintritt ins Rentenalter & libertarian & social & 0 & 0 \\
innere Sicherheit & staatliche Überwachung & libertarian & social & 0 & 0 \\
neoliberal & marktwirtschaftlich orientierte Positionen & right & economic & 0 & 0 \\
populistisch & bürgernah & libertarian & social & 0 & 0 \\
schlecht & förderungswürdig & libertarian & social & 0 & 0 \\
schwul & homosexuell & libertarian & social & 0 & 0 \\
\midrule
\multicolumn{6}{c}{\textbf{Generated Word Pairs}} \\
\midrule
Abschiebung & Rückführung & libertarian & social & 0 & 0 \\
Abwrackprämie & Verschrottungsprämie & right & economic & 0 & 1 \\
Agrarindustrie & Landwirtschaft & right & economic & 0 & 0 \\
Almosen & Staatliche Unterstützung & left & economic & 1 & 0 \\
Altersarmut & Rente & right & economic & 0 & 0 \\
Altersarmut & Ruhestand mit Einschränkungen & no shift & economic & 0 & 0 \\
Altersheim & Seniorenresidenz & no shift & social & 0 & 0 \\
Altersvorsorge & Ruhestandsplanung & no shift & economic & 0 & 0 \\
Arbeitskampf & Streik & right & economic & 0 & 0 \\
Arbeitslosengeld & Übergangsunterstützung & left & economic & 0 & 0 \\
Arbeitslosigkeit & Erwerbsunterbrechung & left & economic & 0 & 0 \\
Arbeitslosigkeit & Übergangsphase & left & economic & 0 & 0 \\
Arbeitsmarkt & Beschäftigungsmarkt & left & economic & 0 & 0 \\
Arbeitsplatzverlust & Berufliche Neuorientierung & no shift & economic & 0 & 0 \\
Arbeitsplatzverlust & Jobvernichtung & no shift & economic & 0 & 1 \\
Arbeitssuchende & Arbeitsmuffel & right & economic & 0 & 1 \\
Armut & Einkommensschwäche & left & economic & 0 & 0 \\
Armutsbekämpfung & Sozialer Ausgleich & left & economic & 0 & 0 \\
Asylbewerber & Wirtschaftsflüchtlinge & right & economic & 0 & 0 \\
Asylsuchende & Wirtschaftsflüchtlinge & right & economic & 0 & 0 \\
Atomkraft & Saubere Energie & right & economic & 0 & 0 \\
Atomkraft & Strahlenschleudern & left & economic & 0 & 1 \\
Atomkraftwerk & Energiepark & no shift & economic & 0 & 0 \\
Aufnahmeeinrichtung & Flüchtlingslager & authoritarian & social & 0 & 0 \\
Automobilindustrie & Mobilitätssektor & right & economic & 0 & 0 \\
Bildungssystem & Bildungsfabrik & libertarian & social & 0 & 1 \\
Blechlawine & Straßenverkehr & right & economic & 1 & 0 \\
Bundeswehr & Kriegstreiber & libertarian & social & 0 & 1 \\
Bürokratie & Papierkrieg & right & economic & 0 & 1 \\
Chancengleichheit & Bildungsgerechtigkeit & authoritarian & social & 0 & 0 \\
Chancengleichheit & Gleichstellung & authoritarian & social & 0 & 0 \\
Datenkrake & Digitalisierung & right & economic & 1 & 0 \\
Datenspeicherung & Informationsarchivierung & no shift & economic & 0 & 0 \\
Digitalisierung & Zukunftstechnologien & right & economic & 0 & 0 \\
Diktatur & Gewaltherrschaft & no shift & social & 0 & 0 \\
Diplomatie & internationale Zusammenarbeit & left & economic & 0 & 0 \\
Dreckschleudern & Kohlekraftwerke & right & economic & 1 & 0 \\
Ehe für alle & Gleichgeschlechtliche Ehe & authoritarian & social & 0 & 0 \\
Eingliederung & Integration & authoritarian & social & 0 & 0 \\
Einstiegslöhne & Niedriglohnsektor & left & economic & 0 & 0 \\
Elite & Elfenbeinturm & left & economic & 0 & 1 \\
Energiekonzerne & Strommonopolisten & left & economic & 0 & 0 \\
Energiekosten & Stromwucher & left & economic & 0 & 1 \\
Energiepolitik & Energiemix & no shift & economic & 0 & 0 \\
Entlassung & Freisetzung & right & economic & 0 & 0 \\
Entwicklungshilfe & Almosenhilfe & right & economic & 0 & 1 \\
Europäische Union & Zwangskorsett & right & economic & 0 & 1 \\
Fachkräftemangel & Arbeitskräftebedarf & no shift & economic & 0 & 0 \\
Fake News & Falschinformation & libertarian & social & 0 & 0 \\
Familienpolitik & Bevormundungspolitik & authoritarian & social & 0 & 1 \\
Faschismus & Nationalismus & authoritarian & social & 0 & 0 \\
Finanzhilfe & Unterstützungsleistung & no shift & economic & 0 & 0 \\
Finanztransaktionssteuer & Börsenbremse & right & economic & 0 & 1 \\
Flugverkehr & Luftmobilität & no shift & economic & 0 & 0 \\
Flüchtlinge & Schutzsuchende & libertarian & social & 0 & 0 \\
Flüchtlingshilfe & Asylantenflut & right & economic & 0 & 1 \\
Flüchtlingskontingent & Einwanderungssteuerung & no shift & economic & 0 & 0 \\
Flüchtlingskrise & Asylantenschwemme & right & economic & 0 & 1 \\
Flüchtlingsunterkunft & Asyllager & right & economic & 0 & 0 \\
Fortschritt & Wirtschaftswachstum & right & economic & 0 & 0 \\
Freihandel & Marktöffnung & no shift & economic & 0 & 0 \\
Freihandel & Raubtierkapitalismus & left & economic & 0 & 1 \\
Freihandelspolitik & Marktöffnungspolitik & no shift & economic & 0 & 0 \\
Fremde & Immigranten & no shift & social & 0 & 0 \\
Friedensmission & Krieg & libertarian & social & 0 & 0 \\
Friedensmission & Militärischer Einsatz & libertarian & social & 0 & 0 \\
Gefängnis & Justizvollzugsanstalt & authoritarian & social & 0 & 0 \\
Gentechnik & Frankenfood & left & economic & 0 & 1 \\
Geringverdiener & Leistungsschwach & right & economic & 0 & 1 \\
Geschwindigkeitsbegrenzung & Sicherheitsregelung & libertarian & social & 0 & 0 \\
Gesundheitssystem & Krankenmühle & right & economic & 0 & 1 \\
Gesundheitsversorgung & Gesundheitswesen & no shift & none & 0 & 0 \\
Gesundheitsversorgung & Medizinische Betreuung & no shift & economic & 0 & 0 \\
Globalisierung & Wirtschaftskolonialismus & left & economic & 0 & 0 \\
Grundeinkommen & Faulenzergehalt & right & economic & 0 & 1 \\
Grundrechte & Menschenrechte & no shift & social & 0 & 0 \\
Grundversorgung & Existenzsicherung & left & economic & 0 & 0 \\
Hartz IV & Sozialhilfe & left & economic & 0 & 0 \\
Haushaltsdisziplin & Sparpolitik & left & economic & 0 & 0 \\
Hochschulpolitik & Akademisierung & authoritarian & social & 0 & 0 \\
Homophobie & Ablehnung von Homosexuellen & libertarian & social & 0 & 0 \\
Homophobie & Schwulenhass & libertarian & social & 0 & 0 \\
Importzölle & Schutzmaßnahmen für den Markt & left & economic & 0 & 0 \\
Innere Sicherheit & Polizeistaat & libertarian & social & 0 & 0 \\
Kapitalismus & Ausbeutungssystem & left & economic & 0 & 1 \\
Kapitalismus & Turbokapitalismus & left & economic & 0 & 0 \\
Kapitalismuskritik & Wirtschaftsreform & right & economic & 0 & 0 \\
Kinderarbeit & Sklavenarbeit & left & economic & 0 & 0 \\
Kinderbetreuung & Verwahranstalt & libertarian & social & 0 & 0 \\
Klassenkampf & Reichensteuer & right & economic & 0 & 0 \\
Klimaanpassung & Umweltgestaltung & authoritarian & social & 0 & 0 \\
Klimakiller & Umweltsünder & right & economic & 1 & 1 \\
Klimakrise & Klimahysterie & right & economic & 0 & 1 \\
Klimaschutz & Umweltschutz & no shift & economic & 0 & 0 \\
Klimaschutz & Umweltschutzmaßnahmen & right & economic & 0 & 0 \\
Klimaschutz & Ökodiktatur & authoritarian & social & 0 & 1 \\
Klimawandel & Wetteranomalien & authoritarian & social & 0 & 0 \\
Konsolidierung & Schuldenabbau & right & economic & 0 & 0 \\
Konsolidierung & Sparpolitik & left & economic & 0 & 0 \\
Korruption & Lobbyismus & right & economic & 0 & 0 \\
Korruption & Staatsdiebstahl & left & economic & 0 & 0 \\
Krankenkassenbeiträge & Gesundheitskostenbeteiligung & right & economic & 0 & 0 \\
Krieg & Konfliktmanagement & authoritarian & social & 0 & 0 \\
Kriegsführung & Schutzmaßnahme & libertarian & social & 0 & 0 \\
Krisenbewältigung & Krisenmanagement & no shift & social & 0 & 0 \\
Kündigung & Trennung im Arbeitsverhältnis & right & economic & 0 & 0 \\
Lobbyismus & Bestechungspolitik & left & economic & 0 & 1 \\
Lobbyismus & Interessensvertretung & right & economic & 0 & 0 \\
Lobbyismus & Interessenvertretung & right & economic & 0 & 0 \\
Lohngleichheit & Einkommensgerechtigkeit & left & economic & 0 & 0 \\
Lohnkürzungen & Gehaltsanpassungen & right & economic & 0 & 0 \\
Lügenpresse & Massenmedien & libertarian & social & 1 & 0 \\
Marionettenregierung & Regierung & libertarian & social & 1 & 0 \\
Marktwirtschaft & Kapitalistisches System & left & economic & 0 & 0 \\
Massentierhaltung & Intensivlandwirtschaft & right & economic & 0 & 0 \\
Massenvernichtungswaffen & Vernichtungsmaschinen & no shift & social & 0 & 0 \\
Meinungsfreiheit & Hassrede & authoritarian & social & 0 & 0 \\
Menschenrechtsverletzung & Zivilrechtliche Verletzung & authoritarian & social & 0 & 0 \\
Migration & Kulturimport & authoritarian & social & 0 & 0 \\
Migration & Wanderung & libertarian & social & 0 & 0 \\
Migration & Zuwanderung & libertarian & social & 0 & 0 \\
Migrationspolitik & Grenzverrücktheit & right & none & 0 & 1 \\
Migrationspolitik & Willkommenskultur & libertarian & social & 0 & 0 \\
Mitbestimmung & Partizipation & no shift & social & 0 & 0 \\
Monopolbildung & Unternehmensfusion & right & economic & 0 & 0 \\
Nachhaltigkeit & Zukunftsfähigkeit & no shift & economic & 0 & 0 \\
Nachhaltigkeit & Ökowahn & right & economic & 0 & 1 \\
Naturkatastrophen & Umweltereignisse & right & economic & 0 & 0 \\
Niedrigzinsphase & Finanzierungsvorteil & no shift & economic & 0 & 0 \\
Niedrigzinspolitik & Wachstumsförderung & right & economic & 0 & 0 \\
Obdachlose & Penner & authoritarian & social & 0 & 1 \\
Obdachlosigkeit & Wohnungslosigkeit & no shift & social & 0 & 0 \\
Pflegenotstand & Personalmangel & right & economic & 0 & 0 \\
Polizeigewalt & Sicherheitsmaßnahme & authoritarian & social & 0 & 0 \\
Polizeigewalt & Staatsterror & libertarian & social & 0 & 1 \\
Polizeikontrollen & Schikane & libertarian & social & 0 & 1 \\
Preiserhöhung & Tarifangleichung & right & economic & 0 & 0 \\
Privatschule & Bildung für die Elite & libertarian & social & 0 & 0 \\
Propaganda & Öffentlichkeitsarbeit & authoritarian & social & 0 & 0 \\
Protest & Bürgerengagement & libertarian & social & 0 & 0 \\
Protest & Chaotenaufstand & authoritarian & social & 0 & 1 \\
Protestbewegung & Berufsdemonstranten & authoritarian & social & 0 & 1 \\
Protestbewegung & Chaotenbewegung & authoritarian & social & 0 & 1 \\
Rassismus & Fremdenfeindlichkeit & authoritarian & social & 0 & 0 \\
Reichtum & Wohlstand & no shift & economic & 0 & 0 \\
Rentenkürzung & Rentenkollaps & left & economic & 0 & 0 \\
Rentenkürzung & Rentenraub & left & economic & 0 & 1 \\
Rentenlücke & Vorsorgebedarf & right & economic & 0 & 0 \\
Rentenpolitik & Altersversorgung & left & economic & 0 & 0 \\
Rentenreform & Altersarmutspolitik & left & economic & 0 & 0 \\
Rentenreform & Alterssicherung & no shift & economic & 0 & 0 \\
Rüstungsausgaben & Verteidigungsausgaben & no shift & economic & 0 & 0 \\
Rüstungsexporte & Waffenhandel & left & economic & 0 & 0 \\
Rüstungspolitik & Verteidigungspolitik & no shift & economic & 0 & 0 \\
Schmutzkampagne & Wahlkampf & no shift & social & 0 & 0 \\
Schulabbrecher & Bildungsabgänger & libertarian & social & 0 & 0 \\
Schulden & Investitionen in die Zukunft & left & economic & 0 & 0 \\
Schuldenbremse & Haushaltsdisziplin & left & economic & 0 & 0 \\
Sicherheitsmaßnahme & Überwachung & libertarian & social & 0 & 0 \\
Solidarität & Zwangssolidarität & authoritarian & social & 0 & 1 \\
Solidaritätszuschlag & Strafsteuer & right & economic & 0 & 1 \\
Sozialabgaben & Beitragspflichten & right & economic & 0 & 0 \\
Sozialabgaben & Zwangsabgaben & right & economic & 0 & 1 \\
Sozialhilfe & Existenzsicherung & left & economic & 0 & 0 \\
Sozialleistungen & Sozialschmarotzerei & right & economic & 0 & 1 \\
Sozialleistungen & Staatliche Unterstützung & left & economic & 0 & 0 \\
Sozialstaat & Umverteilungsstaat & left & economic & 0 & 0 \\
Sozialversicherung & Beitragslast & right & economic & 0 & 0 \\
Sozialwohnungen & Ghettos & right & economic & 0 & 1 \\
Staatsfunk & Öffentlich-Rechtliche Medien & libertarian & social & 0 & 0 \\
Stadtentwicklung & Stadtgestaltung & libertarian & social & 0 & 0 \\
Sterbehilfe & Lebensendeberatung & libertarian & social & 0 & 0 \\
Steuerbetrug & Steuerflucht & right & economic & 0 & 0 \\
Steuererhöhung & Einnahmenanpassung & left & economic & 0 & 0 \\
Steuererhöhung & Steueranpassung & right & economic & 0 & 0 \\
Steuererleichterung & Steuerprivilegien & left & economic & 0 & 0 \\
Steuerflucht & Steuervermeidung & right & economic & 0 & 0 \\
Steuerlast & Steuerabzocke & right & economic & 0 & 1 \\
Steueroase & Steuerparadies & no shift & economic & 0 & 0 \\
Steuerpolitik & Einnahmenmanagement & right & economic & 0 & 0 \\
Steuerreform & Steueranpassung & no shift & economic & 0 & 0 \\
Straßenbau & Betonpolitik & left & economic & 0 & 0 \\
Straßenbau & Infrastrukturentwicklung & no shift & economic & 0 & 0 \\
Studiengebühren & Bildungsbremse & left & economic & 0 & 1 \\
Subventionen & Marktverzerrung & right & economic & 0 & 0 \\
Subventionen & Wirtschaftsförderung & no shift & economic & 0 & 0 \\
Subventionen & Zukunftsinvestitionen & left & economic & 0 & 0 \\
Todesstrafe & Ultimative Strafe & authoritarian & social & 0 & 0 \\
Umweltschutz & Umweltfanatismus & right & economic & 0 & 1 \\
Umweltschutz & Öko-Terrorismus & authoritarian & social & 0 & 1 \\
Umweltverschmutzung & Naturverlust & left & economic & 0 & 0 \\
Umweltverschmutzung & ökologische Belastung & right & economic & 0 & 0 \\
Ungleichheit & soziale Schieflage & left & economic & 0 & 0 \\
Unternehmenssteuer & Wettbewerbsbremse & right & economic & 0 & 0 \\
Verkehrsinfrastruktur & Mobilitätsnetzwerk & no shift & economic & 0 & 0 \\
Verkehrsstau & Hohe Verkehrsdichte & no shift & economic & 0 & 0 \\
Verkehrswende & Mobilitätschaos & right & economic & 0 & 1 \\
Vorruhestand & Frühpensionierung & no shift & economic & 0 & 0 \\
Waffenexport & Blutgeschäfte & left & economic & 0 & 1 \\
Waffenexport & Verteidigungslieferung & right & economic & 0 & 0 \\
Waffenlieferungen & Verteidigungshilfe & right & economic & 0 & 0 \\
Wahlbetrug & Wahlfälschung & no shift & social & 0 & 0 \\
Wahlrechtsreform & Wahlrechtsanpassung & no shift & social & 0 & 0 \\
Wahlversprechen & Zukunftsvisionen & no shift & social & 0 & 0 \\
Wirtschaftswachstum & Wachstumswahn & left & economic & 0 & 1 \\
Wohngemeinschaft & Studentenbude & authoritarian & social & 0 & 0 \\
Wohnungsbau & Bau von Lebensräumen & left & economic & 0 & 0 \\
Wohnungsbau & Wohnraumschaffung & left & economic & 0 & 0 \\
Wutbürger & Bürgerprotest & libertarian & social & 1 & 0 \\
Zensur & Informationskontrolle & authoritarian & social & 0 & 0 \\
Zivildienst & Zwangsdienst & libertarian & social & 0 & 1 \\
Zivilgesellschaft & Gutmenschenverein & authoritarian & social & 0 & 1 \\
Zuwanderung & Überfremdung & authoritarian & social & 0 & 0 \\
Übergewicht & Wohlstandskilos & no shift & social & 0 & 0 \\
Überwachungsgesetz & Orwell-Gesetz & libertarian & social & 0 & 1 \\
Überwachungsgesetz & Sicherheitsgesetz & authoritarian & social & 0 & 0 \\
Überwachungsstaat & Sicherheitspolitik & authoritarian & social & 0 & 0 \\

\bottomrule
\caption{Word pair annotations.
We annotate the political shift of the word pair, the axis on which this shift occurs and whether the first or second word is judgmental. 
For the complete annotation guideline see Appendix \ref{sec:appendix_guideline}.}
\label{tab:wordpair_shifts}
\end{longtable}

\subsection{Annotation Guideline}
\label{sec:appendix_guideline}

\textbf{Main Question}:
Compared to the first word, are people who use the second word in the same context more politically left or right positioned?

Definitions by Petrik 2010 and construct by Kitschelt 1994:

\begin{itemize}
    \item Economic axis:
        \begin{itemize}
        \item \textbf{Equality (left)}:
        “view that assets should be redistributed by a cooperative collective agency (the state, in socialist tradition or a network of communes, in the libertarian or anarchist tradition).”
        \item \textbf{Liberty (right)}:
        “view that the economy should be left to the market system, to voluntary competing individuals and organizations.”
        \end{itemize}

    \item Social axis:
        \begin{itemize}
        \item \textbf{Libertarianism}: 
        “the idea that personal freedom as well as voluntary and equal participation should be maximized. This would be the full realization of liberty and equality in a democratic sense. Parts of that view are ideas like autonomous, direct democratic institutions beyond state and market, transformation of gender roles, enjoyment and self-determination over traditional and religious order.”
        \item \textbf{Authoritarianism}: 
        “the belief that authority and religious or secular traditions should be complied with. Equal participation and a free choice of personal behavior are rejected as being against human nature or against necessary hierarchies for a stable society.”
        \end{itemize}
\end{itemize}

For each word pair, follow these steps:
\begin{enumerate}
    \item \textbf{Identify Topic} \\
    When people use these words, are they referring to economic topics, social topics? Choose the topic which fits better. If you don’t know, which sometimes happens if the words are too similar, choose “I don’t know”.

    \item \textbf{If Economic: Assign Economic Shift}
    \begin{itemize}
        \item Leftward shift (toward economic equality): If the second word increases state intervention, wealth redistribution, or social welfare (e.g., “Besteuerung” → “bedingungsloses Einkommen”), label it as a shift toward the economic left.
        \item Rightward shift (toward market liberty): If the second word reduces state intervention and emphasizes free-market mechanisms or privatization (e.g., “gesetzliche Krankenversicherung” → “private Versicherung”), label it as a shift toward the economic right.
    \end{itemize}

    \item \textbf{If Social: Assign Social Shift}
    \begin{itemize}
        \item Libertarian shift (toward individual freedom \& inclusivity): If the second word implies increased personal freedom, reduced state control, or more progressive values (e.g., “Zensur” → “freie Meinungsäußerung”), label it as a libertarian shift.
        \item Authoritarian shift (toward order \& tradition): If the second word reinforces hierarchy, state control, or traditional values (e.g. “Ehe für Alle” → “traditionelle Familie“), label it as a shift toward authoritarianism.
    \end{itemize}

    \item \textbf{Assign Judgmental} \\
    If the first word includes a negative or positive judgment (e.g. “Marionettenregierung”, “Schmutzkampagne”), click “first word judgmental”. This is about semantic judgment, where no prior knowledge is required (e.g. “Sparpolitik” is meant negatively but is not obviously/semantically negative). \\
    If the second word includes a negative or positive judgment (e.g. “Marionettenregierung”), click “second word judgmental”. This is about semantic judgment, where no prior knowledge is required (e.g. “Sparpolitik” is meant negatively but is not obviously/semantically negative).
\end{enumerate}

Examples:
\small
\begin{table}[h]
    \centering
    {\small
    \begin{tabular}{@{}llllc@{}}
        \toprule
        \textbf{First Word} & \textbf{Second Word} & \textbf{Dimension} & \textbf{Shift} & \textbf{Judgement} \\ 
        \midrule
        Streik & Arbeitskampf & economic & Left & 0 0 \\ 
        Asylbewerber & Wirtschaftsflüchtling & social & Libertarian & 0 0 \\ 
        Dreckschleudern & Kohlekraftwerk & economic & Right & 1 0 \\ 
    \bottomrule
    \end{tabular}
    }
\end{table}

\subsection{Inter-Annotator Agreements}

The five annotators consist of a student, a PhD student in computer science and a PhD student in political science and a PhD student in computer science with background in political science.
All annotators are native German speakers.
All annotators participated willingly without compensation with knowledge of potential publication of anonymized annotations and meta-information. The inter-annotator agreements were calculated with Krippendorff's alpha and can be found in Table \ref{tab:inter-annotator_agreements}.

\small
\begin{table}[h]
    \centering
    {\small
    \begin{tabular}{lc} %
        \toprule
        \textbf{Variable} & \textbf{Krippendorff's alpha} \\ 
        \midrule
        axis & 0.336 \\ 
        shift & 0.224 \\ 
        base word judgmental & 0.243 \\ 
        exchange word judgmental & 0.495 \\ 
    \bottomrule
    \end{tabular}

    \caption{Inter-Annotator Agreements of the word pair annotations. Five annotators assessed the axis (social, economic, NA), the shift (libertarian/left, authoritarian/right, none) and whether the base and exchange words are judgmental or not judgmental.}
    \label{tab:inter-annotator_agreements}
        }
\end{table}

\subsection{Annotators Bias}
\label{sec:appendix_annotator_bias}
\begin{figure}[h]
\centering
  \includegraphics[width=0.5\textwidth]{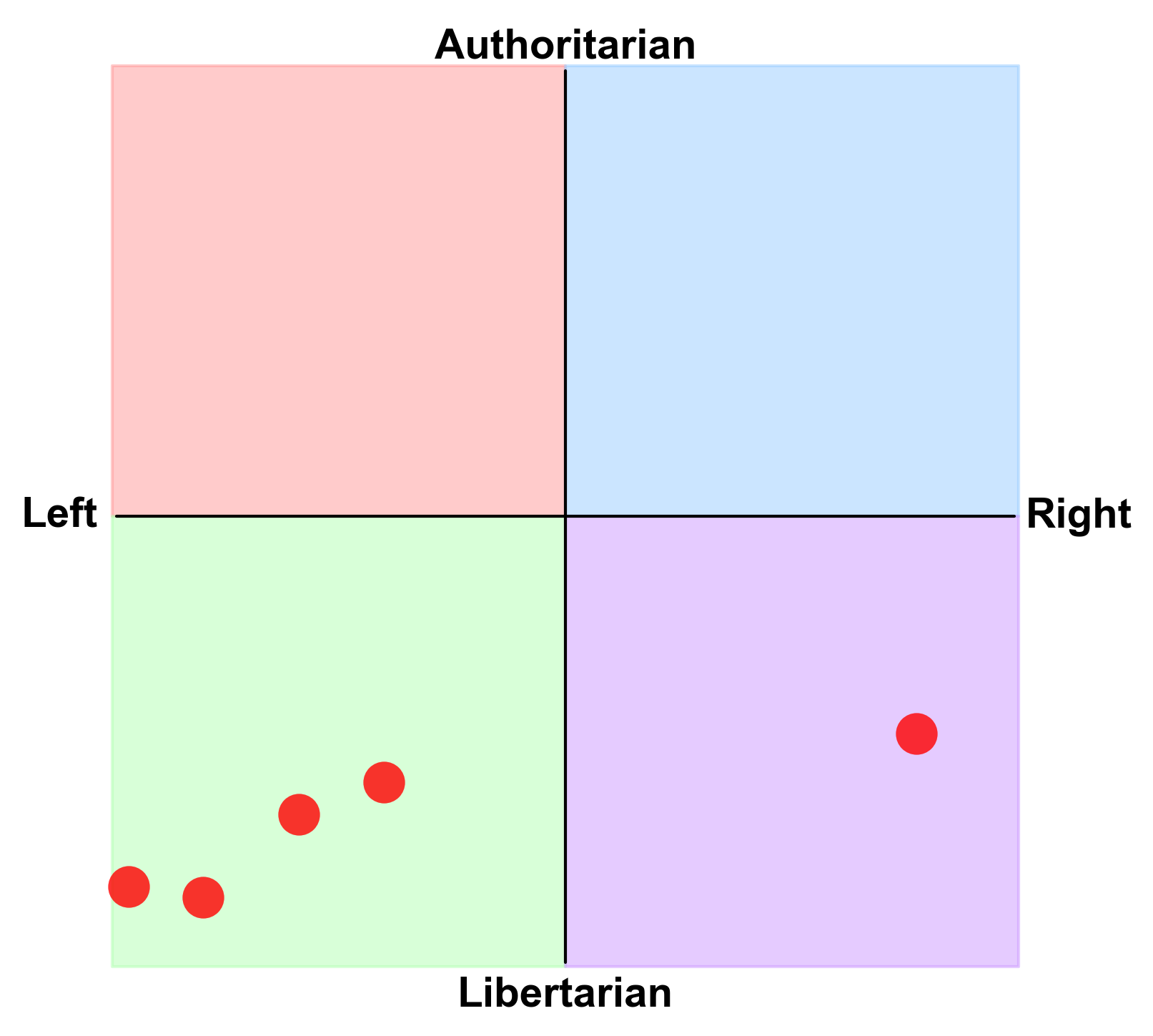}
  \caption{Annotators Political Bias}
\end{figure}

\section{Technical Setup}
\label{sec:technical_setup}

We use statsmodels\footnote{\url{https://www.statsmodels.org/stable/index.html}} for statistical modeling.
Furthermore, we used Cursor\footnote{\url{https://www.cursor.com}} for 
the construction of diagrams and ChatGPT\footnote{\url{https://chatgpt.com/}} to assist with a few specific formulations in the paper.

\end{document}